\documentclass[letterpaper]{article} % DO NOT CHANGE THIS
\usepackage{aaai2026}  % DO NOT CHANGE THIS
\usepackage{times}  % DO NOT CHANGE THIS
\usepackage{helvet}  % DO NOT CHANGE THIS
\usepackage{courier}  % DO NOT CHANGE THIS
\usepackage[hyphens]{url}  % DO NOT CHANGE THIS
\usepackage{graphicx} % DO NOT CHANGE THIS
\usepackage{natbib}  % DO NOT CHANGE THIS AND DO NOT ADD ANY OPTIONS TO IT
\usepackage{caption} % DO NOT CHANGE THIS AND DO NOT ADD ANY OPTIONS TO IT
\usepackage{algorithm}
\usepackage{algorithmic}
\usepackage{booktabs}
\usepackage{multirow}
\usepackage{longtable}
\usepackage{newfloat}
\usepackage{listings}
\DeclareCaptionStyle{ruled}{labelfont=normalfont,labelsep=colon,strut=off} % DO NOT CHANGE THIS
\floatstyle{ruled}
\newfloat{listing}{tb}{lst}{}
\floatname{listing}{Listing}
\title{MedCAL-Bench: A Comprehensive Benchmark on Cold-Start Active Learning with Foundation Models for Medical Image Analysis}
\author{
    Ning Zhu\textsuperscript{\rm 1}\equalcontrib,
    Xiaochuan Ma\textsuperscript{\rm 2}\equalcontrib,
    Shaoting Zhang\textsuperscript{\rm 2,3},
    Guotai Wang\textsuperscript{\rm 2,3}\thanks{Corresponding Author},
}

\affiliations{
    \textsuperscript{\rm 1}Glasgow College, University of Electronic Science and Technology of China, Chengdu, China\\

    \textsuperscript{\rm 2}School of Mechanical and Electrical Engineering, University of Electronic Science and Technology of China, Chengdu, China\\

    \textsuperscript{\rm 3}Shanghai AI Lab, Shanghai, China
}

\usepackage{bibentry}
\begin{document}

\maketitle

\begin{abstract}
Cold-Start Active Learning (CSAL) aims to select informative samples for annotation without prior knowledge, which is important for improving annotation efficiency and model performance under a limited annotation budget in medical image analysis. Most existing CSAL methods rely on Self-Supervised Learning (SSL) on the target dataset for feature extraction, which is inefficient and limited by insufficient feature representation. Recently, pre-trained Foundation Models (FMs) have shown powerful feature extraction ability with a potential for better CSAL. However, this paradigm has been rarely investigated, with a lack of benchmarks for comparison of FMs in CSAL tasks. To this end, we propose \textbf{MedCAL-Bench}, the first systematic FM-based CSAL benchmark for medical image analysis. We evaluate 14 FMs and 7 CSAL strategies across 7 datasets under different annotation budgets, covering classification and segmentation tasks from diverse medical modalities. It is also the first CSAL benchmark that evaluates both the feature extraction and sample selection stages.  Our experimental results reveal that: 1) Most FMs are effective feature extractors for CSAL, with DINO family performing the best in segmentation; 2) The performance differences of these FMs are large in segmentation tasks, while small for classification; 3) Different sample selection strategies should be considered in CSAL on different datasets, with Active Learning by Processing Surprisal (ALPS) performing the best in segmentation while RepDiv leading for classification. The code is available at \underline{\url{https://github.com/HiLab-git/MedCAL-Bench}}.
\end{abstract}

%     \link{Code}{to be filled}
%     \link{Datasets}{to be filled}
%     \link{Extended version}{to be filled}

\section{Introduction}
Deep learning has shown promising performance in various medical image analysis tasks, such as organ or tumor segmentation~\cite{rokuss2025lesionlocator} and disease classification~\cite{Phan2024decomposing}. However, its success relies highly on numerous annotated training samples, which creates a significant barrier in scenarios where labeled data is scarce and annotating large datasets is infeasible~\cite{jin2025label,wang2024comprehensive,kumari2023data,Zhong2025UniSAL}. \textbf{Active Learning} (AL) has emerged as an effective solution by selecting the most informative samples for labeling to minimize the annotation cost. Traditional \textit{warm-start active learning} relies on a labeled subset that is randomly selected to iteratively refine the model by selecting valuable samples for further annotation~\cite{Luo2024Uncertainty,Wei2024BaSAL}. However, the randomness in selecting initial labeled data often leads to sub-optimal model performance. 

To address this issue, \textbf{Cold-Start Active Learning} (CSAL) has been proposed to better select the initial subset of images for annotation, which not only can be used for the initialization of traditional multi-round AL, but is also appealing when a single round of sample selection from an unannotated dataset is required~\cite{Jin2022Oneshot}. While most traditional AL methods leverage uncertainty and diversity for sample selection~\cite{ma2025sugfw,doucet2025bridging,hacohen2022active}, uncertainty-based methods are hardly applicable for CSAL, as there are no models that have been adapted to the target unannotated dataset for reliable uncertainty estimation. Instead, most CSAL methods use diversity for sample selection, where a feature extractor is firstly trained by \textbf{Self-Supervised Learning} (SSL) on the unannotated dataset~\cite{zhu2025csal}, and then used to project the samples into the feature space for diversity-based selection. However, these methods suffer from limited feature representation ability, due to the small size of target dataset for SSL, and the bias of features learned by the pretext task. 
 
\begin{figure*}[h]
  \centering
  \includegraphics[width=0.87\textwidth]{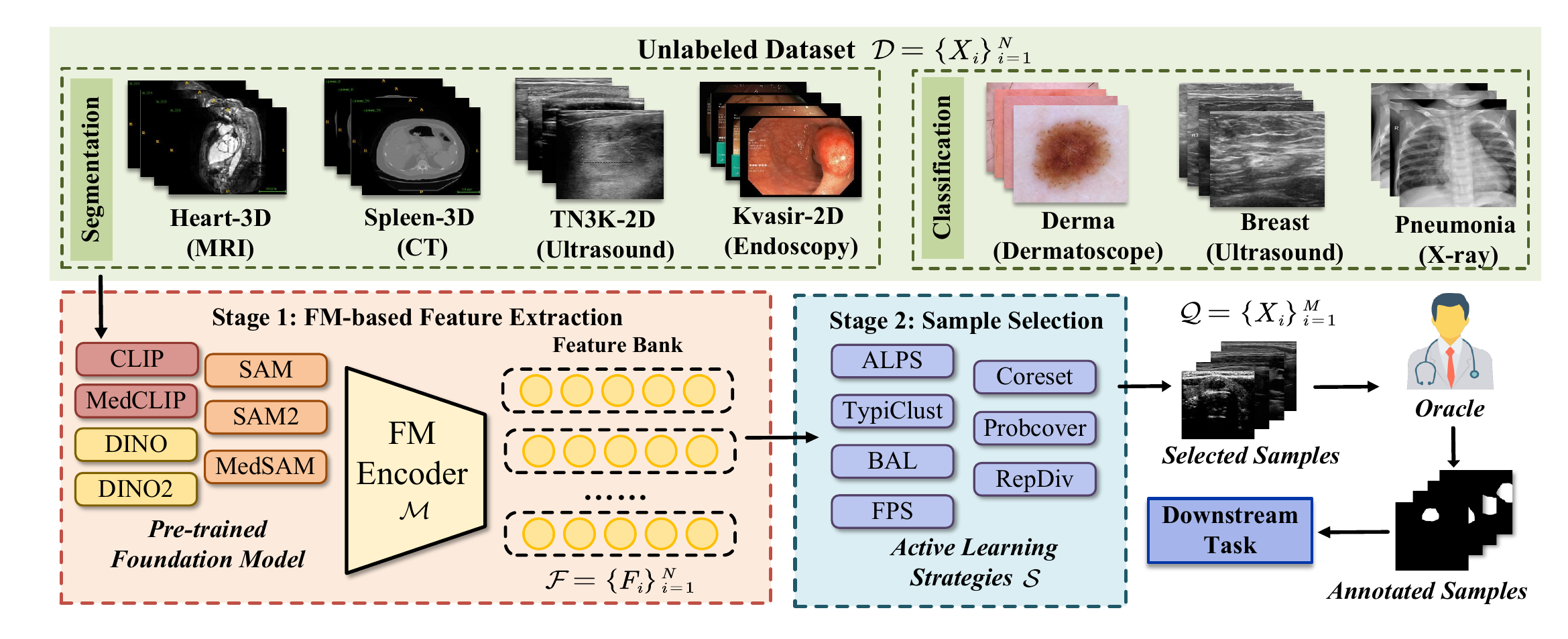}
  \caption{The workflow of \textbf{MedCAL-Bench} that provides a comprehensive evaluation of both feature extractors and sample selection methods of CSAL on 7 medical datasets for both segmentation and classification tasks. Specifically, we evaluate the effectiveness of foundation models for feature extraction in CSAL, which has not been investigated in existing benchmarks.} 
  \label{fig:structure}
\end{figure*}

In recent years, \textbf{Foundation Models} (FMs) have demonstrated exceptional capabilities in extracting semantic and transferable features from images. Pre-trained on large scale and diverse datasets, FMs like CLIP~\cite{radford2021learning}, SAM~\cite{kirillov2023segment} and DINO~\cite{zhang2022dino} have significantly advanced the state of visual representation learning, with generalizable feature representations for unseen image types in downstream datasets, which provides a new type of feature extractors for CSAL. Despite huge potential, FM-based CSAL has been scarcely investigated in current works, and there is also a lack of benchmarks for evaluation of both FMs and sample selection strategies in CSAL. Current works on AL benchmarks mainly focus on warm-start scenarios~\cite{Zhan2021comparative,Holzmüller2023framework,Werner2024cross-domain} with only little attention paid to CSAL~\cite{Liu2023COLosSAL,chen2022making}. Moreover, these works often neglect feature extraction methods and only concentrate on comparison of sample selection strategies. 

To this end, we propose a novel paradigm for CSAL by leveraging pre-trained FMs for feature extraction. Unlike SSL-based CSAL, this approach avoids training the feature extractor for each downstream dataset, which is data- and computation-efficient for sample selection in CSAL.  In addition, to comprehensively understand the effectiveness, generalizability and robustness of FMs in CSAL, we present \textbf{MedCAL-Bench}, the first comprehensive benchmark for evaluating pre-trained FM-based CSAL in medical image analysis. We evaluate 14 variants of FMs and 7 CSAL strategies on 7 medical image datasets for segmentation and classification from 6 modalities. Our main contributions are:
\begin{itemize}
    \item We propose a novel CSAL paradigm that leverages pre-trained FMs instead of dataset-specific SSL with powerful feature representation for better sample selection.
    \item We introduce \textbf{MedCAL-Bench}: the first FM-based CSAL benchmark in medical image applications that evaluates 14 FMs and 7 CSAL strategies across 7 datasets under different annotation budget levels.
    \item We present the first comprehensive analysis on the effect of both feature extractor and sample selection rule on CSAL in medical image analysis.  
  \end{itemize}  
Our major findings include: 1) Most FM-driven CSAL methods outperform random selection on the majority of datasets; 2) DINO models prove to be the most effective feature extractors for CSAL in segmentation; 3) Surprisingly, medical FMs do not demonstrate superiority compared with general FMs; 4) The performance of CSAL methods is highly sensitive to FM feature extractor choice; 5)  No single CSAL method can consistently achieve top performance across all datasets, and Active Learning by Processing Surprisal (ALPS) performs the best for segmentation, while RepDiv leads in classification. 

\section{Related Work}
\subsubsection{Cold-Start Active Learning.}
CSAL can be divided into two categories: uncertainty-based and diversity-based approaches. Typical uncertainty-based approaches, such as ProxyRank-Ent~\cite{Nath2021diminishing} and ProxyRank-Var~\cite{Yang2017Suggestive}, utilize Hounsfield Unit (HU) intensity window %~\cite{nath2022warm}  
or Otsu thresholding~\cite{Liu2023COLosSAL} to quantify uncertainty estimation and select samples with higher uncertainties. However, these methods require prior knowledge of the specific dataset with limited generalizability, and prove to perform worse than diversity-based counterparts~\cite{doucet2025bridging,hacohen2022active}. Diversity-based approaches aim to select samples from diverse regions to cover the data distribution. Common methods include Coreset~\cite{sener2017active},  ALPS~\cite{yuan2020cold}, CALR~\cite{jin2022cold}, FPS~\cite{Jin2022Oneshot}, Typiclust~\cite{hacohen2022active}, Probcover~\cite{Yehuda2022Probcover},  Unsupervised Selective Labeling (USL)~\cite{Wang2022Unsupervised}, RepDiv~\cite{Shen2020deep} and BAL~\cite{Li2024BAL}. 

\subsubsection{Foundation Models.}
Recent advances in FMs have demonstrated remarkable capabilities in visual representation learning. The Segment Anything Model (SAM)~\cite{kirillov2023segment,ma2024segment,ravi2024sam}, introduces a promptable segmentation framework that leverages a powerful vision transformer backbone to achieve zero-shot generalization across various segmentation tasks. In contrast, CLIP~\cite{radford2021learning,wang2022medclip} pioneers contrastive language-image pre-training, aligning visual and textual embeddings through a dual-encoder architecture. Moreover, DINO~\cite{zhang2022dino} and its variants~\cite{oquab2023dinov2, jose2024dinov2} employ knowledge distillation with vision transformers to learn semantically rich representations without explicit labels. 

\subsubsection{Active Learning Benchmark.}
Existing works on AL benchmarks are mostly tailored for warm-start scenarios. For instance, ~\citet{Zhan2021comparative} compared pool-based AL strategies, and \citet{Holzmüller2023framework} provided a more systematic benchmark for AL in regression tasks. \citet{Werner2024cross-domain} extends AL benchmark for cross-domain task evaluation. Despite these encouraging works, benchmarks for CSAL in medical image analysis are very limited. Specifically, \citet{Liu2023COLosSAL} presented a simple CSAL baseline for 3D medical image segmentation, while \citet{chen2022making} focused on benchmarking CSAL methods for classification. However, these benchmarks only focus on the comparison of sample selection strategies, ignoring the effect of feature extractor on CSAL.

\section{Methodology}

\subsubsection{Problem Setting.} The overall framework of our proposed \textbf{MedCAL-Bench} is illustrated in \textbf{Fig.~\ref{fig:structure}}.
Considering an unlabeled training set $\mathcal{D}=\{X_i\}_{i=1}^N$, we aim to select a subset $\mathcal{Q}=\{X_i\}_{i=1}^{M}$ for labeling using an FM $\mathcal{M}$ and a feature-based sample selection method $\mathcal{S}$. Firstly, by projecting each training image $X_i \in \mathcal{D}$ into the feature space via  $\mathcal{M}$, a feature bank $\mathcal{F} = \{F_i\}_{i=1}^N$ is constructed: 
\begin{equation}
    \mathcal{F} = \{\mathcal{M}(X_i) | X_{i} \in \mathcal{D}\}
\end{equation}
Subsequently, given an annotation budget $M$ ($M \ll N$), a query set $\mathcal{Q}$ with a size of $M$ is extracted by a feature-based sample selection method $\mathcal{S}$, which is formulated as:

\begin{equation}
    \mathcal{Q} = \mathcal{S}(\mathcal{F})
\end{equation}
Finally, the query set $\mathcal{Q}$ is annotated by the oracle to train a task-specific model (i.e. segmentation and classification) using supervised learning. 

\subsubsection{Foundation Models.} Recent FMs are mostly based on Transformer architectures, and each of them comes with a pre-trained feature extractor. To analyze the effect of these pre-trained feature extractors on CSAL, we consider the following 14 FMs in three main categories: 
\begin{enumerate}
    \item \textbf{5 SAM-like Models}: SAM~\cite{kirillov2023segment} is an interactive segmentation model pre-trained on a large-scale annotated RGB image dataset with over 1 billion masks. MedSAM~\cite{ma2024segment} adapts SAM to medical tasks through fine-tuning with annotated medical images. SAM2~\cite{ravi2024sam} further extends SAM’s segmentation capabilities to video by adding multi-scale features, hybrid supervision, and uncertainty modeling. We used the \textit{ViT-B} and \textit{ViT-H} versions of SAM, the \textit{ViT-B} version of MedSAM, as well as the \textit{2b+} and \textit{2.1b+} versions of SAM2 for comparison.

    \item \textbf{5 CLIP-like Models}: 
    CLIP~\cite{radford2021learning} is a vision-language FM pre-trained on 400 million image-text pairs for zero-shot classification through cross-modal alignment. MedCLIP~\cite{wang2022medclip} employs domain-specific contrastive learning with medical reports and anatomy-aware tokenization to improve embedding relevance for clinical tasks. We included \textit{RN50x64}, \textit{ViT-L\_14}, and \textit{ViT-L\_14@336px} versions of CLIP, and the \textit{ResNet} and \textit{ViT} versions of MedCLIP for comparison.
    
    \item \textbf{4 DINO-like Models}: 
    DINO~\cite{zhang2022dino} is a self-supervised vision transformer trained on large-scale unannotated images  via knowledge distillation with multi-crop augmentation. Its successor DINOv2~\cite{oquab2023dinov2} scales this paradigm to larger models (e.g., ViT-G/14) using curated data pipelines and  teacher-student synchronization. In this work, we included 4 DINO-based feature extractors: \textit{ViT-S\_16} and \textit{ViT-B\_16}, and the DINOv2 models \textit{ViT-B\_14} and \textit{ViT-G\_14}.

\end{enumerate} 
In addition, to analyze the superiority of FMs over traditional pre-trained feature extractors, we compare these FMs with a  ResNet18 model~\cite{He2016deep} pre-trained on ImageNet-1k~\cite{Deng2009ImageNet} via supervised learning.

\subsubsection{Sample Selection Strategies.}
With the features extracted by the above FMs, we apply the following 7 typical and recent sample selection rules to obtain images for annotation: 1) \textbf{ALPS}~\cite{yuan2020cold} uses $K$-Means to cluster features in $\mathcal{F}$ into $M$ clusters. In each cluster, the sample closest to the centroid is selected. 2) \textbf{Typiclust}~\cite{hacohen2022active} also uses $K$-Means to obtain $M$ clusters, and selects the sample with the highest typicality, defined as the inverse of the average Euclidean distance to the nearest neighbors. 3) \textbf{BAL}~\cite{Li2024BAL} selects the sample with the lowest Cluster Distance Difference (CCD) from each cluster after $K$-Means clustering, where CCD is measured by the distance between the first and second nearest centroids. 4) \textbf{FPS}~\cite{Jin2022Oneshot} selects a first cluster center randomly, and then iteratively selects new centers with probability inversely proportional to their distance from existing centers, followed by Farthest Point Sampling strategy~\cite{moenning2003fast}. 5) \textbf{Coreset}~\cite{sener2017active} constructs a weighted subset that approximates the original dataset with a minimized distribution error. 6) \textbf{Probcover}~\cite{Yehuda2022Probcover} maximizes data coverage by selecting informative samples using a covering framework. It forms a directed graph and selects the sample with the highest out-degree, and updates coverage after each selection. 7) \textbf{RepDiv}~\cite{Shen2020deep} uses a greedy strategy to balance representativeness (similarity to unlabeled samples) and diversity (dissimilarity to selected samples) within the selected samples.

\begin{table}[h]
\caption{Dataset summary. Slice numbers are listed for the dataset size of Heart (20 volumes) and Spleen (41 volumes).}
\centering
\resizebox{\linewidth}{!}{
\begin{tabular}{ccccccc}
\toprule
 \textbf{Dataset} & \textbf{Dim.} & \textbf{Modality} & \textbf{Task} & \textbf{Size} & \textbf{Classes} & \textbf{Budgets}\\
\midrule

 Heart & 3D & MRI  & \multirow{4}{*}{Seg} & 1,567 & 2 & 5\%, 10\%, 15\% \\
 Spleen & 3D & CT &  & 3,650 & 2 & 2\%, 4\%, 10\%\\
 Kvasir & 2D & Endoscopy &  & 1,000 & 2 & 10\%, 15\%, 20\%\\
 TN3K & 2D & Ultrasound &  & 3,493 & 2 & 5\%, 7.5\%, 10\%\\
\midrule
 Derma & 2D & Dermatoscope & \multirow{3}{*}{Cls} & 10,015 & 7 & 1\%, 2\%, 3\%\\
 Breast & 2D & Ultrasound &  & 800 & 2 & 5\%, 10\%, 15\%\\
 Pneumonia & 2D & X-ray &  & 5,856 & 2 & 1\%, 2\%, 3\%\\
 \bottomrule
\end{tabular}}
\label{tab:datasets}
\end{table}

\begin{figure*}[h]
  \centering
  \includegraphics[width=0.91\textwidth]{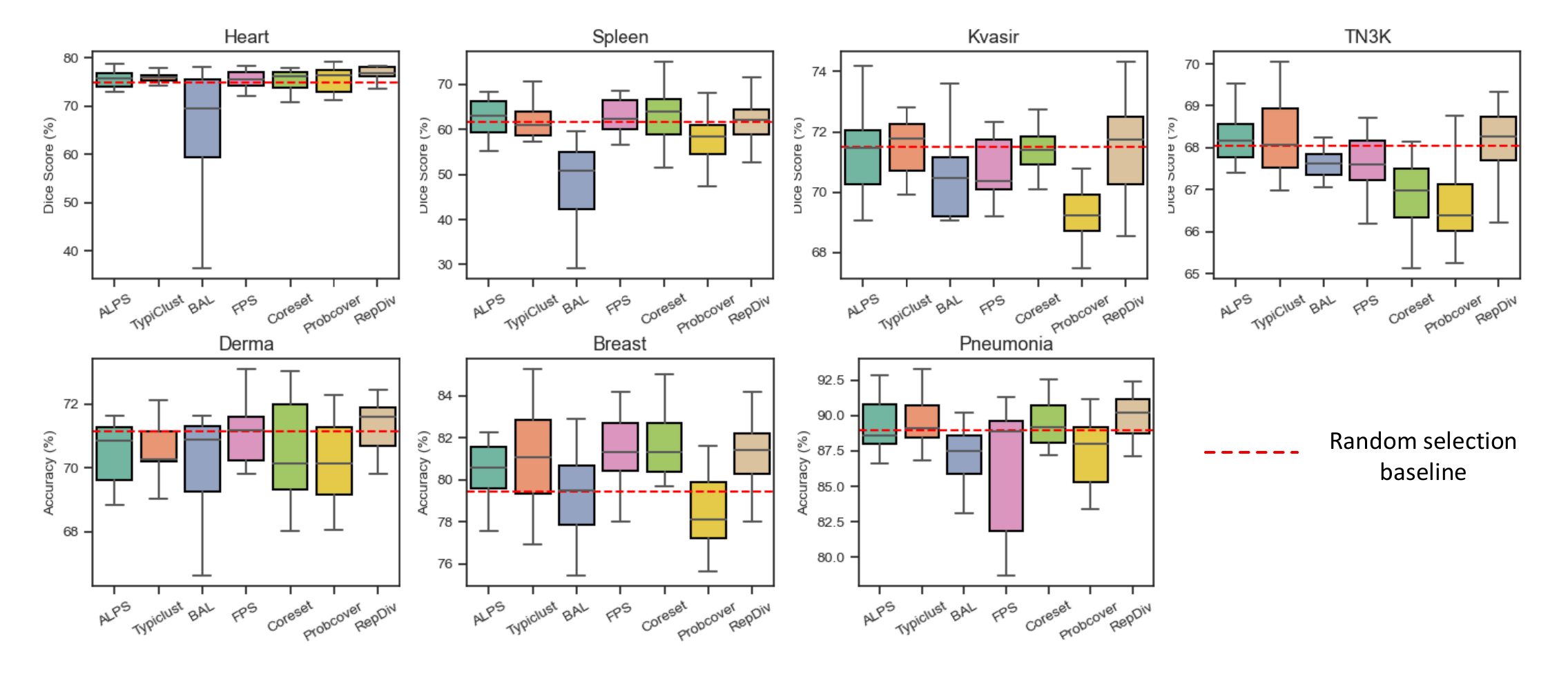}
  \caption{Performance overview of CSAL methods compared with random selection. For each sample selection method on each dataset, the boxplot shows distribution of Dice/accuracy when using different feature extractors (i.e., the 14 FMs). } 
  \label{fig:generalization}
\end{figure*}

\section{Experiment}

\subsubsection{Datasets.}\label{subsec:dataset}
We concentrate on two common medical image analysis tasks: segmentation and classification. For segmentation, we used Heart and Spleen datasets from Medical Segmentation Decathlon~\cite{simpson2019large} for 3D segmentation from MRI/CT images. For 2D segmentation, we used Kvasir~\cite{Pogorelov2017KVASIR} and TN3K~\cite{gong2021multi-task,gong2022thyroid} datasets for segmentation of lesions from endoscopic and ultrasound images, respectively. For classification tasks, we used three datasets (Derma, Breast, Pneumonia) from the MedMNIST dataset~\cite{Yang2023MedMNIST}. The detailed information about the modality, dataset size and class number is shown in \textbf{Table~\ref{tab:datasets}}. For  Kvasir and classification datasets, we followed the official data split provided by the organizers.  For Spleen and TN3K datasets, we adopted a 7:1:2 ratio for training, validation, and test sets. For Heart dataset, a 5:3:2 split was used due to the small number of volumes.

\subsubsection{Implementation Details.}\label{subsec:implementation_details}
For segmentation tasks, each dataset was processed by re-sampling to the average spacing in the training dataset. The image intensity was normalized by z-score normalization  using the mean and standard deviation. As the FMs are based on 2D feature extractors, for 3D datasets, we used slice-level feature extraction, sample selection  and segmentation in the experiments, and 2D nnUNet~\cite{isensee2021NnUNet} was used for training and inference. The queried training set was augmented by Gaussian noise injection, Gaussian blurring, brightness enhancement, contrast adjustment, gamma transformation, and randomized spatial transformations. We used SGD with an initial learning rate of 0.01, momentum of 0.9, weight decay of 3$\times 10^{-5}$, and a cosine annealing schedule. The batch size was 8 and the epoch number was 1000. For classification tasks, data augmentation consisted of color jitter, random flip, random rotation, and random scaling. We used ResNet18~\cite{He2016deep} pre-trained on ImageNet-1k~\cite{Deng2009ImageNet} as the classification model. It was optimized with AdamW~\cite{loshchilov2019decoupled}, using an initial learning rate of 0.001, betas of (0.9, 0.999), and weight decay of 0.05, with a cosine annealing schedule. The batch size was 128 and the epoch number was 100. As shown in \textbf{Table~\ref{tab:datasets}},  we considered three different annotation budgets on each dataset for robust evaluation, and the performance was measured by Dice score and accuracy for segmentation and classification tasks, respectively. Experiments were conducted in PyTorch on a Linux server with four NVIDIA GeForce GTX 1080Ti GPUs.

\subsection{Performance Overview of CSAL with FMs} 
We first analyze the overall performance of CSAL methods with different FM feature extractors. For each sample selection method, we calculated the mean Dice/accuracy (averaged across three annotation budgets) obtained by each of the 14 FMs, and the distribution of these 14 results is shown by the boxplot in \textbf{Fig.~\ref{fig:generalization}}. The CSAL methods were compared with a baseline method of random sample selection (averaged by 5 different random seeds). It can be observed that on each dataset, random selection was competitive and performed better than some CSAL methods. For the segmentation tasks, ALPS generally outperformed the other methods, and for classification tasks, RepDiv obtained the best performance in general. The large variance in the boxplots show that these methods are sensitive to feature extractors. Except for BAL, all the sample selection methods may outperform random selection on all the datasets when a proper feature extractor is employed, indicating the large performance gap between different FMs for feature extraction. 

\begin{table*}[h]
\caption{Comparison between different FMs for 3D segmentation in Dice (\%), with the best in \textbf{bold} and the second-best \underline{underlined}. SS: Sample Selection method.}
\centering
\resizebox{\textwidth}{!}{

\begin{tabular}{ccccc|cccc|cccc|cccc}
\toprule
 & \multicolumn{8}{c|}{Heart (3D)} %(top-3 SS methods: ALPS, FPS, Typiclust)}
 & \multicolumn{8}{c}{Spleen (3D)} % (top-3 SS methods: ALPS, FPS, RepDiv)} 
 \\
\midrule
\multirow{2}{*}{\textbf{Foundation Model}} & \multicolumn{4}{c|}{\textbf{Averaged on 7 SS methods}} & \multicolumn{4}{c|}{\textbf{Averaged on top-3 SS methods}} & \multicolumn{4}{c|}{\textbf{Averaged on 7 SS methods}} & \multicolumn{4}{c}{\textbf{Averaged on top-3 SS methods}}  \\

& 5\% & 10\% & 15\% & Avg & 5\% & 10\% & 15\% & Avg & 2\% & 4\% & 10\% & Avg  & 2\% & 4\% & 10\% & Avg \\ \midrule

    ResNet18 & 71.85 & 78.63 & 79.33 & 76.60 & 73.53 & \textbf{80.01} & 80.03 & \textbf{77.86} & 53.19 & 54.07 & 66.20 & 57.82 & 51.46 & 55.60 & 68.04 & 58.37 \\

    SAM-ViT-B & 59.36 & 72.52 & 78.92 & 70.27 & 67.79 & 78.49 & 80.06 & 75.45 & 53.90 & 61.52 & 63.65 & 59.69 & \textbf{60.98} & 66.56 & 64.48 & 64.00  \\
    SAM-ViT-H & 58.88 & 69.08 & 75.65 & 67.87 & 65.70 & 77.80 & 78.68 & 74.06 & 48.51 & 62.05 & 61.80 & 57.45 & 45.58 & 64.86 & 56.55 & 55.66 \\
    MedSAM-ViT-B & 64.36 & 73.43 & 77.14 & 71.64 & 65.47 & 77.34 & 79.89 & 74.24 & 54.72 & 57.58 & 60.96 & 57.75 & 58.74 & 62.42 & 67.37 & 62.85 \\
    SAM2-2b+ & 68.14 & 77.19 & \textbf{80.67} & 75.33 & 69.74 & 77.56 & \underline{80.94} & 76.08 & \underline{59.47} & \textbf{65.90} & 67.17 & \textbf{64.18} & 60.35 & \textbf{73.37} & \underline{70.87} & \textbf{68.20} \\
    SAM2-2.1b+ & 65.06 & 74.52 & 79.01 & 72.86 & 70.45 & 76.45 & \textbf{81.04} & 75.98 & \textbf{59.83} & 62.01 & \textbf{68.89} & \underline{63.58} & \underline{60.57} & 62.77 & \textbf{71.96} & 65.10 \\
    CLIP-RN50x64 & 66.72 & 77.18 & 80.13 & 74.67 & 64.91 & 77.19 & 79.75 & 73.95 & 53.48 & 59.62 & 62.66 & 58.59 & 57.29 & 60.52 & 64.80 & 60.87 \\
    CLIP-ViT-L\_14 & 62.57 & 76.22 & 79.42 & 72.74 & 68.92 & 76.41 & 79.38 & 74.91 & 51.04 & \underline{64.16} & \underline{67.37} & 60.86 & 55.28 & 66.98 & 70.98 & 64.41 \\
    CLIP-ViT-L\_14@336px & 60.53 & 76.16 & 80.02 & 70.24 & 70.58 & 76.56 & 79.31 & 75.48 & 50.57 & 60.55 & 65.02 & 58.71 & 52.78 & 60.95 & 66.92 & 60.22 \\
    MedCLIP-ResNet & \underline{73.67} & 78.51 & \underline{80.16} & \underline{77.45} & 72.55 & 76.69 & 79.85 & 74.91 & 51.64 & 59.33 & 62.84 & 57.94 & 51.09 & 60.44 & 65.23 & 58.92 \\
    MedCLIP-ViT & \textbf{74.39} & \textbf{78.94} & 79.18 & \textbf{77.50} & \textbf{74.25} & 79.05 & 79.75 & \underline{77.68} & 52.52 & 62.03 & 65.49 & 60.01 & 49.84 & 62.16 & 65.68 & 59.23 \\
    DINO-ViT-S\_16 & 70.73 & \underline{78.67} & 78.35 & 75.92 & 73.51 & \underline{79.85} & 78.75 & 77.37 & 57.14 & 61.00 & 65.09 & 61.08 & 58.97 & 65.74 & 66.85 & 63.85 \\
    DINO-ViT-B\_16 & 69.65 & 77.43 & 79.18 & 75.42 & \underline{73.93} & 79.28 & 79.50 & 77.67 & 58.85 & 62.72 & 66.66 & 62.74 & 60.50 & \underline{68.47} & \underline{70.87} & \underline{66.61} \\
    DINOv2-ViT-B\_14 & 72.83 & 78.47 & 79.45 & 76.92 & 73.68 & 78.81 & 80.32 & 77.60 & 45.99 & 55.64 & 59.08 & 53.57 & 59.99 & 57.86 & 62.26 & 60.03 \\
    DINOv2-ViT-G\_14 & 67.56 & 75.04 & 76.65 & 73.08 & 68.84 & 76.11 & 77.58 & 74.18 & 35.08 & 46.79 & 53.88 & 45.25 & 51.53 & 59.56 & 65.93 & 59.01 \\
    \midrule
    SAM-avg & 63.16 & 73.35 & 78.28 & 71.59 & 67.83 & \underline{77.53} & \textbf{80.12} & 75.16 & \textbf{55.29} & \textbf{61.81} & \underline{64.49} & \textbf{60.53} & \underline{57.24} & \textbf{66.00} & 66.24 & \textbf{63.16}  \\
    CLIP-avg & \underline{67.58} & \textbf{77.40} & \textbf{79.78} & \underline{74.52} & \underline{70.24} & 77.18 & \underline{79.61} & \underline{75.39} & \underline{51.85} & \underline{61.14} & \textbf{64.68} & \underline{59.22} & 53.25 & 62.21 & \textbf{66.72} & 60.73 \\
    DINO-avg & \textbf{70.19} & \textbf{77.40} & \underline{78.41} & \textbf{75.34} & \textbf{72.49} & \textbf{78.51} & 79.04 & \textbf{76.71} & 49.27 & 56.54 & 61.18 & 57.95 & \textbf{57.75} & \underline{62.91} & \underline{66.48} & \underline{62.38} \\ \bottomrule
\end{tabular}}
\label{tab:foundation_models_heart_spleen}
\end{table*}

\begin{table*}[h]
\caption{Comparison between  FMs for 2D segmentation in Dice (\%). 
SS: Sample Selection method.}

\centering
\resizebox{\textwidth}{!}{

\begin{tabular}{ccccc|cccc|cccc|cccc}
\toprule
 & \multicolumn{8}{c|}{Kvasir (2D)} % (top-3 SS methods: ALPS, Typiclust, RepDiv)} 
 & \multicolumn{8}{c}{TN3K (2D)} % (top-3 SS methods: ALPS, Typiclust, RepDiv)}
 \\
\midrule
\multirow{2}{*}{\textbf{Foundation Model}} & \multicolumn{4}{c|}{\textbf{Averaged on 7 SS methods}} & \multicolumn{4}{c|}{\textbf{Averaged on top-3 SS methods}} & \multicolumn{4}{c|}{\textbf{Averaged on 7 SS methods}} & \multicolumn{4}{c}{\textbf{Averaged on top-3 SS methods}} \\

& 10\% & 15\% & 20\% & Avg & 10\% & 15\% & 20\% & Avg & 5\% & 7.5\% & 10\% & Avg & 5\% & 7.5\% & 10\% & Avg \\ \midrule

    ResNet18 & 67.35 & 72.49 & \textbf{76.06} & \textbf{71.97} & 68.28 & 71.95 & 75.43 & 71.89 & 61.37 & 67.12 & 70.14 & 66.21 & 63.60 & 67.86 & 71.07 & 67.51 \\
    SAM-ViT-B & 66.89 & 71.37 & 74.91 & 71.06 & \textbf{68.52} & 71.14 & 74.50 & 71.39 & 64.14 & 68.19 & 70.23 & 67.52 & 64.12 & 69.50 & 70.76 & 68.13 \\
    SAM-ViT-H & 66.38 & 71.60 & 73.97 & 70.65 & 67.81 & 71.81 & 73.36 & 70.99 & \underline{64.54} & 68.92 & 70.18 & 67.88 & 65.44 & 69.31 & 69.95 & 68.23 \\
    MedSAM-ViT-B & 66.10 & 71.27 & 74.89 & 70.72 & 67.66 & 72.35 & 74.78 & 71.60 & 64.29 & 68.31 & 70.82 & 67.81 & 65.36 & 68.56 & 71.10 & 68.34 \\
    SAM2-2b+ & 65.99 & \underline{72.67} & 74.46 & 71.04 & 66.61 & 70.96 & 74.32 & 70.63 & 64.28 & 68.53 & 70.50 & 67.77 & 64.76 & \underline{69.95} & 71.31 & 68.67 \\
    SAM2-2.1b+ & 65.91 & 71.30 & 73.91 & 70.38 & 66.61 & 70.96 & 74.32 & 70.63 & 64.11 & 68.71 & \underline{70.94} & 67.92 & 64.56 & \textbf{70.20} & 71.32 & 68.70 \\
    CLIP-RN50x64 & 65.40 & 71.06 & 73.90 & 70.12 & 66.41 & 71.47 & 73.85 & 70.58 & 64.18 & \underline{68.95} & \textbf{71.49} & \textbf{68.21} & 64.73 & 68.74 & \underline{71.77} & 68.41 \\
    CLIP-ViT-L\_14 & 67.29 & 72.31 & 75.03 & 71.54 & \underline{68.51} & 73.25 & \underline{75.96} & \underline{72.58}  & 64.15 & 67.63 & 70.34 & 67.37 & 65.12 & 68.11 & 70.34 & 67.86 \\
    CLIP-ViT-L\_14@336px & 66.95 & 72.27 & 74.05 & 71.09 & 65.13 & \underline{73.81} & 74.38 & 71.11 & 63.49 & 68.19 & 70.31 & 67.33 & 64.17 & 68.63 & 71.06 & 67.95 \\
    MedCLIP-ResNet & 63.23 & 70.20 & 74.36 & 69.27 & 62.05 & 71.38 & 74.62 & 69.35 & 64.41 & 67.56 & 70.16 & 67.38 & 64.97 & 66.95 & 71.08 & 67.67 \\
    MedCLIP-ViT & 67.07 & \textbf{72.70} & 75.41 & \underline{71.73} & 67.66 & 72.35 & 74.78 & 71.60 & 63.02 & 67.23 & 70.09 & 66.78 & 63.08 & 67.69 & 70.10 & 66.96 \\
    DINO-ViT-S\_16 & 66.72 & 72.36 & 75.23 & 71.44 & 67.47 & 72.91 & \underline{75.96} & 72.11 & 64.09 & 68.55 & 70.64 & 67.76 & 64.02 & 69.20 & 71.14 & 68.12 \\
    DINO-ViT-B\_16 & 65.11 & 71.92 & 74.69 & 70.57 & 68.10 & \textbf{74.10} & 75.64 & \textbf{72.61} & \textbf{64.83} & \textbf{69.03} & 70.64 & \underline{68.16} & \textbf{66.49} & 69.62 & 71.31 & \textbf{69.14 }\\
    DINOv2-ViT-B\_14 & \textbf{67.62} & 71.89 & \underline{75.51} & 71.67 & 68.00 & 71.77 & \textbf{77.30} & 72.36 & \underline{64.54} & 68.47 & 70.65 & 67.89 & \underline{65.47} & 69.10 & \textbf{71.98} & \underline{68.85} \\
    DINOv2-ViT-G\_14 & \underline{67.40} & 70.95 & 73.80 & 70.72 & 67.26 & 71.88 & 75.75 & 71.63 & 63.51 & 67.73 & 69.68 & 66.97 & 64.50 & 68.66 & 70.30 & 67.82 \\
    \midrule
    SAM-avg & \underline{66.25} & 71.64 & 74.43 & \underline{70.77} & \underline{67.52} & 71.57 & 74.51 & \underline{71.20} & \textbf{64.27} & \textbf{68.53} & \textbf{70.53} & \textbf{67.78} & \underline{64.85} & \textbf{69.50} & \underline{70.89} & \underline{68.41} \\
    CLIP-avg & 65.99 & \underline{71.71} & \underline{74.55} & 70.75 & 65.95 & \underline{72.45} & \underline{74.72} & 71.04 & 63.85 & 67.91 & \underline{70.48} & 67.41 & 64.41 & 68.02 & 70.87 & 67.77  \\
    DINO-avg & \textbf{66.71} & \textbf{71.78} & \textbf{74.81} & \textbf{71.10} & \textbf{67.71} & \textbf{72.67} & \textbf{76.16} & \textbf{72.18} & \underline{64.24} & \underline{68.45} & 70.40 & \underline{67.70} & \textbf{65.12} & \underline{69.15} & \textbf{71.18} & \textbf{68.48} \\ \bottomrule

\end{tabular}}
\label{tab:foundation_models_kvasir_TN3K}
\end{table*}

\begin{table*}[htbp]
  \centering
  \caption{Comparative results of Sample Selection (SS) methods for segmentation in Dice (\%).}
  {\setlength{\tabcolsep}{3.5pt}
  \renewcommand{\arraystretch}{1.05}
 \fontsize{7.5pt}{7.5pt}\selectfont

\begin{tabular}{cccccc|cccc|cccc|cccc}
    \toprule
    
    Foundation & SS & \multicolumn{4}{c|}{Heart (3D)}    & \multicolumn{4}{c|}{Spleen (3D)}   & \multicolumn{4}{c|}{Kvasir (2D)}   & \multicolumn{4}{c}{TN3K (2D)} \\
    
    Model & Method & 5\%    & 10\%    & 15\%   & Avg   & 2\%    & 4\%   & 10\%   & Avg   & 10\%    & 15\%   & 20\%   & Avg   & 5\%   & 7.5\%   & 10\%   & Avg \\
    
    \midrule
    
    \multirow{7}[1]{*}{SAM} & ALPS  & 68.08 & 77.49 & 80.48 & 75.35 & 53.22 & \textbf{68.34} & 67.35 & 62.97 & 66.69 & 71.71 & \underline{75.03} & 71.14 & 64.07 & 69.29 & \textbf{71.80} & 68.39 \\
          & Coreset & 68.19 & 75.47 & \textbf{80.84} & 74.83 & \textbf{66.47} & \underline{66.26} & \textbf{72.23} & \textbf{68.32} & 66.58 & 71.33 & \textbf{75.85} & \underline{71.25} & 63.14 & 67.54 & 69.77 & 66.82 \\
          & FPS   & 67.21 & 77.17 & 79.40 & 74.59 & \underline{59.74} & 66.07 & 65.33 & \underline{63.72} & 67.35 & 71.97 & 73.23 & 70.85 & 63.71 & 67.91 & 69.51 & 67.04 \\
          & Probcover & \underline{71.12} & \underline{77.51} & \underline{80.58} & \textbf{76.40} & 50.11 & 56.11 & 55.56 & 53.93 & 63.11 & 71.05 & 73.45 & 69.20 & 64.67 & 67.34 & 69.52 & 67.18 \\
          & Typiclust & 68.20 & \textbf{77.93} & 80.48 & 75.54 & 58.40 & 63.21 & \underline{69.14} & 63.58 & \underline{67.65} & 70.95 & 74.45 & 71.02 & \underline{65.23} & \textbf{69.75} & 70.56 & \underline{68.51} \\
          & BAL   & 27.29 & 50.83 & 67.53 & 48.55 & 40.29 & 49.12 & 55.80 & 48.40 & 64.20 & \textbf{72.28} & 74.96 & 70.48 & 63.84 & 68.43 & 71.23 & 67.83 \\
          & RepDiv & \textbf{72.04} & 77.01 & 78.64 & \underline{75.89} & 58.76 & 63.58 & 66.06 & 62.80 & \textbf{68.21} & \underline{72.06} & 74.04 & \textbf{71.43} & \textbf{65.24} & \underline{69.47} & \underline{71.35} & \textbf{68.69} \\
    
    \midrule
    
    \multirow{7}[2]{*}{CLIP} & ALPS  & 69.25 & 76.44 & 79.74 & 75.14 & 53.21 & \underline{64.25} & \underline{66.76} & \textbf{61.41} & 66.09 & 71.90 & 74.83 & 70.94 & \textbf{65.00} & 67.38 & 70.61 & 67.66 \\
          & Coreset & 72.12 & \underline{78.10} & \underline{79.84} & \underline{76.69} & 51.00 & 63.12 & 65.30 & 59.81 & \underline{67.10} & 72.24 & \underline{75.02} & \underline{71.45} & 62.37 & 67.95 & 70.22 & 66.85 \\
          & FPS   & 71.37 & 76.57 & 79.82 & 75.92 & \textbf{54.91} & 60.98 & 63.83 & 59.91 & 64.30 & 70.36 & \textbf{75.21} & 69.96 & 64.16 & \textbf{68.76} & 70.59 & \underline{67.83} \\
          & Probcover & 49.68 & 76.23 & \textbf{80.86} & 68.93 & 52.48 & 59.49 & 63.94 & 58.63 & 64.89 & 70.21 & 72.55 & 69.22 & 63.66 & 66.59 & 68.77 & 66.34 \\
          & Typiclust & 69.16 & 78.04 & 79.57 & 75.59 & \underline{53.95} & \textbf{65.14} & 65.01 & \underline{61.37} & 64.75 & \textbf{72.95} & 74.82 & 70.84 & \underline{64.41} & \underline{68.50} & 70.69 & \textbf{67.87} \\
          & BAL   & \textbf{73.15} & 77.62 & 78.92 & 76.56 & 45.75 & 53.59 & 58.32 & 52.55 & \textbf{67.77} & 71.79 & 74.93 & \textbf{71.50} & 63.53 & 68.02 & \textbf{71.26} & 67.60 \\
          & RepDiv & \underline{72.35} & \textbf{78.80} & 79.71 & \textbf{76.96} & 51.64 & 61.41 & \textbf{69.57} & 60.87 & 67.01 & \underline{72.51} & 74.51 & 71.34 & 63.82 & 68.20 & \underline{71.21} & 67.74 \\
    
    \midrule
    
    \multirow{7}[1]{*}{DINO} & ALPS  & 72.62 & \underline{78.77} & 78.74 & \underline{76.71} & \textbf{61.45} & 61.05 & 63.75 & 62.08 & 66.95 & \textbf{73.52} & 75.66 & \underline{72.04} & \textbf{65.49} & \underline{69.29} & \textbf{71.82} & \textbf{68.87} \\
          & Coreset & 68.88 & 77.42 & 79.64 & 75.31 & 37.22 & 49.01 & 57.49 & 47.91 & 66.88 & 71.98 & 74.13 & 71.00 & 63.46 & 67.86 & 69.73 & 67.02 \\
          & FPS   & \underline{72.78} & 78.44 & 78.70 & 76.64 & \underline{57.79} & \textbf{64.41} & 67.25 & \textbf{63.15} & \underline{67.67} & 71.13 & 73.85 & 70.88 & 64.02 & 68.53 & 70.38 & 67.64 \\
          & Probcover & 71.92 & 76.92 & 77.75 & 75.53 & 56.02 & \underline{64.00} & \underline{67.61} & \underline{62.54} & 64.55 & 70.50 & 73.01 & 69.35 & 63.61 & 67.14 & 69.05 & 66.60 \\
          & Typiclust & 72.08 & 78.33 & \underline{79.68} & 76.69 & 52.21 & 56.79 & 63.27 & 57.42 & \textbf{69.39} & \underline{72.36} & \underline{75.95} & \textbf{72.57} & \underline{65.03} & \textbf{69.59} & 70.99 & \underline{68.54} \\
          & BAL   & 58.16 & 72.05 & 74.16 & 68.12 & 26.17 & 37.23 & 40.45 & 34.62 & 64.78 & 70.86 & 74.17 & 69.94 & 63.25 & 68.15 & 69.48 & 66.96 \\
          & RepDiv & \textbf{74.93} & \textbf{79.88} & \textbf{80.19} & \textbf{78.33} & 54.01 & 63.27 & \textbf{68.44} & 61.91 & 66.78 & 72.12 & \textbf{76.88} & 71.93 & 64.85 & 68.56 & \underline{71.35} & 68.25 \\

    \midrule
     \multicolumn{2}{c}{Random selection} & 67.21 & 77.54 & 79.99 & 74.91 & 52.88 & 64.77 & 67.08 & 61.58 & 67.34 & 72.28 & 74.19 & 71.51 & 64.89 & 68.71 & 70.88 & 68.16 \\
    \multicolumn{2}{c}{Fully supervision} & 85.96 & 85.96 & 85.96 & 85.96 & 92.52 & 92.52 & 92.52 & 92.52 & 87.94 & 87.94 & 87.94 & 87.94 & 81.35 & 81.35 & 81.35 & 81.35  \\ 
    \bottomrule

    \end{tabular}}%
\label{tab:csal_compare_segmentation}%
\end{table*}%

\subsection{Evaluation of FM-based CSAL for Segmentation}\label{subsec:seg_results}

\subsubsection{FM Comparison.} For detailed comparison of different FMs for CSAL, we first show the results  on the segmentation datasets in \textbf{Table~\ref{tab:foundation_models_heart_spleen}} and \textbf{Table~\ref{tab:foundation_models_kvasir_TN3K}}. For each FM and each annotation ratio, we report two Dice scores that are averaged across all the 7 sample selection methods and only top-3 sample selection methods, respectively. The top-3 methods are determined based on their average performance for each dataset in \textbf{Fig.~\ref{fig:generalization}}. In addition, we report the average performance for each category of FMs, as shown in the last sections of the Tables.  It shows that DINO series was the most reliable FM for segmentation tasks, with the highest average Dice on Heart (76.71\%) , Kvasir (72.18\%) and TN3K (68.48\%) for the top-3 sample selection methods. Following DINO, SAM and CLIP exhibited comparable feature extraction performance for segmentation tasks, with SAM family showing a slight advantage on Spleen dataset. Moreover, the Dice difference of these three series of FMs was small on 2D datasets ($<$1.0\%), while larger on the 3D datasets ($>$3.8\% on Heart with 7 sample selection methods), showing that the feature extractor is more sensitive on 3D segmentation datasets. In addition, we found except for the Heart dataset (for top-3 sample selection methods) and Kvasir dataset (for 7 sample selection methods), most FMs outperformed ResNet18 on the other datasets.

\subsubsection{Comparison of Sample Selection Methods.} \textbf{Table~\ref{tab:csal_compare_segmentation}} compares the performance of the 7 sample selection methods under different feature extractors, where Dice values obtained by each category  of FMs are averaged. 
When the SAM family was used as the feature extractor, Probcover and Coreset obtained the best performance on the Heart and Spleen datasets, respectively, and RepDiv outperformed the others on the Kvasir and TN3K datasets. RepDiv, ALPS, FPS and Typiclust perform comparably well when the CLIP variants were used as feature extractors. When using DINO family for feature extraction, Typiclust and ALPS obtained the best performance on the Kvasir and TN3K datasets, respectively. In general, RepDiv, ALPS and Typiclust are the top three sample selection methods on different datasets with different feature extractors, and outperformed random selection, showing their effectiveness and robustness. 

As also demonstrated in \textbf{Fig.~\ref{fig:generalization}}, \textbf{Table~\ref{tab:csal_compare_segmentation}} shows that BAL performed significantly worse on the 3D datasets than the other competitors. We found that BAL tends to select many background slices or slices with a small foreground region (please refer to Section 1 of the \textbf{Appendix} for detailed visualization), which can lead to under-segmentation on the 3D datasets. This is mainly because that BAL prefers samples near the cluster boundaries, and 2D slices with a small foreground region can easily be present in such regions in the feature space.

\begin{table*}[t!]
\caption{Comparative results of FMs for classification in accuracy (\%), with the best in \textbf{bold} and the second-best \underline{underlined}.}
\centering
{\renewcommand{\arraystretch}{1.1}
 \fontsize{7.5pt}{7pt}\selectfont

\begin{tabular}{ccccc|cccc|cccc}
\toprule
\multirow{2}{*}{Foundation Model} & \multicolumn{4}{c|}{Derma} & \multicolumn{4}{c|}{Breast} & \multicolumn{4}{c}{Pneumonia} \\
& 1\% & 2\% & 3\% & Avg & 5\% & 10\% & 15\% & Avg & 1\% & 2\% & 3\% & Avg \\ \midrule

    ResNet18 & 67.85 & 69.39 & 73.40 & 70.21 & 77.20 & 78.48 & 80.86 & 78.85 & 86.65 & \underline{90.64} & 88.30 & 88.53 \\

    SAM-ViT-B & 68.71 & 69.98 & 71.98 & 70.22 & 76.83 & \underline{82.23} & 83.70 & 80.92 & 86.93 & \textbf{91.39} & 90.45 & 89.59 \\
    SAM-ViT-H & 66.89 & 69.18 & 72.76 & 69.61 & 78.11 & \textbf{82.33} & 82.05 & 80.83 & 84.66 & 90.25 & 89.03 & 87.98 \\
    MedSAM-ViT-B & 69.57 & 69.74 & 72.96 & 70.76 & 75.73 & 81.59 & \underline{83.88} & 80.40 & 84.48 & 89.42 & 87.00 & 86.97 \\
    SAM2-2b+ & 68.62 & 69.53 & 72.75 & 70.30 & 77.38 & 82.05 & 83.61 & 81.01 & 85.83 & 87.43 & \underline{90.91} & 88.06 \\
    SAM2-2.1b+ & 68.24 & 69.58 & 72.85 & 70.22 & 77.84 & 81.96 & \textbf{84.25} & \underline{81.35} & 86.86 & 89.79 & 89.95 & 88.87 \\
    CLIP-RN50x64 & 68.94 & 69.58 & 73.12 & 70.55 & 76.01 & 81.87 & 83.06 & 80.31 & 84.50 & 87.77 & 87.07 & 86.45 \\
    CLIP-ViT-L\_14 & 69.27 & 70.40 & \underline{73.67} & 71.11 & \textbf{79.67} & 81.77 & 82.88 & \textbf{81.44} & 86.52 & 89.49 & 89.56 & 88.52 \\
    CLIP-ViT-L\_14@336px & \textbf{69.99} & 70.44 & 73.66 & \underline{71.36} & 77.93 & 81.68 & 81.05 & 80.22 & \textbf{89.29} & 89.06 & \textbf{90.96} & \textbf{89.77} \\
    MedCLIP-ResNet & 69.04 & \underline{70.87} & 73.22 & 71.04 & 75.73 & 79.12 & 83.52 & 79.46 & 88.53 & 89.24 & 88.42 & 88.73 \\
    MedCLIP-ViT & 67.79 & 68.70 & 72.50 & 69.66 & 75.09 & 80.40 & 83.52 & 79.67 & 88.10 & 88.76 & 89.45 & 88.77 \\
    DINO-ViT-S\_16 & 68.72 & 69.18 & \textbf{74.53} & 70.81 & \textbf{79.67} & 80.59 & 82.97 & 81.07 & 87.04 & 88.99 & 85.94 & 87.32 \\
    DINO-ViT-B\_16 & \underline{69.65} & \textbf{71.40} & 73.62 & \textbf{71.56} & 77.65 & 82.14 & 83.61 & 81.14 & \underline{89.08} & 90.52 & 89.40 & \underline{89.67} \\
    DINOv2-ViT-B\_14 & 68.53 & 69.93 & 72.83 & 70.43 & 76.28 & 81.50 & 80.31 & 79.37 & 86.06 & 87.64 & 88.83 & 87.51 \\
    DINOv2-ViT-G\_14 & 68.73 & 69.26 & 72.99 & 70.32 & 78.30 & 79.67 & 82.24 & 80.07 & 83.45 & 90.02 & 87.77 & 87.08 \\
    \midrule
    SAM-avg & 68.41 & 69.60 & 72.66 & 70.22 & \underline{77.18} & \textbf{82.03} & \textbf{83.50} & \textbf{80.90} & 85.75 & \textbf{89.66} & \textbf{89.47} & \underline{88.29} \\
    CLIP-avg & \textbf{69.01} & \textbf{70.00} & \underline{73.23} & \underline{70.74} & 76.89 & 80.97 & \underline{82.81} & 80.22 & \textbf{87.39} & 88.86 & \underline{89.09} & \textbf{88.45} \\
    DINO-avg & \underline{68.91} & \underline{69.94} & \textbf{73.49} & \textbf{70.78} & \textbf{77.98} & \underline{80.98} & 82.28 & \underline{80.41} & \underline{86.41} & \underline{89.29} & 87.99 & 87.90 \\

\bottomrule
\end{tabular}
%}
}
\label{tab:foundation_models_combined}
\end{table*}

\begin{table*}[t!]
\caption{Comparative results of Sample Selection (SS) methods for classification in accuracy (\%), with the best in \textbf{bold} and the second-best \underline{underlined}.}
\centering

  {\setlength{\tabcolsep}{4pt}
  \renewcommand{\arraystretch}{1.05}
 \fontsize{7.5pt}{7.5pt}\selectfont
\begin{tabular}{cccccc|cccc|cccc}
\toprule
Foundation & SS  & \multicolumn{4}{c|}{Derma} & \multicolumn{4}{c|}{Breast} & \multicolumn{4}{c}{Pneumonia} \\
 Model & Method & 1\% & 2\% & 3\% & Avg & 5\% & 10\% & 15\% & Avg & 1\% & 2\% & 3\% & Avg \\ \midrule

    \multirow{7}[1]{*}{SAM} & ALPS  & 67.96 & 68.77 & 73.00 & 69.91 & 73.08 & \textbf{85.26} & 83.97 & 80.77 & 86.57 & 90.61 & \textbf{92.40} & \underline{89.86} \\
          & Coreset & \textbf{69.79} & \underline{70.51} & \underline{73.23} & \textbf{71.18} & \underline{79.49} & \underline{83.97} & 78.85 & 80.77 & 85.58 & 89.10 & 89.45 & 88.04 \\
          & FPS   & 68.95 & \textbf{70.66} & \textbf{73.74} & \underline{71.11} & \textbf{82.05} & 83.33 & 84.62 & \textbf{83.33} & 87.88 & \textbf{92.15} & 88.11 & 89.38 \\
          & Probcover & \underline{69.30} & 69.48 & 71.73 & 70.17 & 78.85 & 82.05 & \textbf{87.18} & \underline{82.69} & 76.41 & 81.99 & 88.17 & 82.19 \\
          & Typiclust & 67.63 & 70.07 & 72.30 & 70.00 & 73.08 & 76.92 & 80.77 & 76.92 & \textbf{89.65} & 91.28 & \underline{90.64} & \textbf{90.52} \\
          & BAL   & 66.98 & 68.50 & 72.29 & 69.26 & 76.92 & 78.85 & 83.97 & 79.91 & 85.35 & \underline{91.38} & 90.00 & 88.91 \\
          & RepDiv & 68.22 & 69.21 & 72.35 & 69.92 & 78.21 & \underline{83.97} & \underline{85.90} & \underline{82.69} & \underline{88.82} & 91.09 & 87.50 & 89.14 \\
    \midrule
    \multirow{7}[2]{*}{CLIP} & ALPS  & 69.04 & \textbf{70.81} & 73.28 & \textbf{71.04} & 78.21 & 76.28 & \underline{85.90} & 80.13 & 88.98 & \textbf{89.91} & 87.92 & 88.93 \\
          & Coreset & \underline{69.43} & 69.81 & 73.46 & 70.90 & \underline{79.49} & 82.05 & \textbf{88.46} & \textbf{83.33} & 88.08 & 88.08 & 86.28 & 87.48 \\
          & FPS   & 68.51 & 69.80 & \textbf{73.68} & 70.66 & \textbf{80.77} & 82.05 & 83.33 & \underline{82.05} & \textbf{89.84} & \underline{89.81} & \textbf{92.92} & \textbf{90.85} \\
          & Probcover & 68.30 & 69.75 & 72.94 & 70.33 & 71.79 & 78.21 & 76.28 & 75.43 & 83.53 & 86.80 & \underline{89.93} & 86.75 \\
          & Typiclust & \textbf{69.75} & 69.45 & 72.83 & 70.68 & 75.64 & 80.77 & 84.62 & 80.34 & 88.59 & 89.65 & 89.33 & 89.19 \\
          & BAL   & 68.74 & \underline{70.61} & \underline{73.66} & \underline{71.00} & 77.56 & \underline{83.33} & 81.41 & 80.77 & 83.14 & 88.30 & 87.63 & 86.36 \\
          & RepDiv & 69.27 & 69.75 & 72.82 & 70.61 & 62.18 & 80.13 & 84.62 & 75.64 & \underline{89.55} & 89.52 & 89.62 & \underline{89.56} \\
    \midrule
    \multirow{7}[1]{*}{DINO} & ALPS  & 68.49 & 70.00 & 73.27 & 70.59 & 77.56 & \underline{84.62} & \underline{84.62} & 82.27 & \underline{88.06} & \textbf{91.99} & 86.74 & 88.93 \\
          & Coreset & \textbf{69.90} & 69.39 & \underline{74.41} & \underline{71.23} & 78.85 & \underline{84.62} & \underline{84.62} & \underline{82.70} & 82.17 & 83.46 & 82.53 & 82.72 \\
          & FPS   & 69.43 & \textbf{70.84} & \textbf{74.57} & \textbf{71.61} & 77.56 & 80.77 & 83.97 & 80.77 & 87.46 & \underline{91.43} & \textbf{90.10} & \underline{89.66} \\
          & Probcover & 68.23 & 69.86 & 71.95 & 70.01 & 77.56 & 82.69 & 82.05 & 80.77 & 87.86 & 89.78 & 89.07 & 88.90 \\
          & Typiclust & \underline{69.72} & 69.66 & 74.12 & 71.16 & \underline{79.49} & 81.41 & \underline{84.62} & 81.84 & 86.58 & 90.15 & 89.22 & 88.65 \\
          & BAL   & 68.06 & 69.50 & 72.98 & 70.18 & 71.79 & 75.64 & 79.49 & 75.64 & 83.65 & 87.10 & 88.90 & 86.55 \\
          & RepDiv & 68.54 & \underline{70.34} & 73.17 & 70.68 & \textbf{80.77} & \textbf{85.26} & \textbf{85.90} & \textbf{83.98} & \textbf{89.06} & 91.15 & \underline{89.35} & \textbf{89.85} \\

    \midrule
    \multicolumn{2}{c}{Random selection}  & 69.58 & 70.76 & 73.10 & 71.15 & 73.72 & 80.13 & 84.49 & 79.45 & 85.61 & 90.58 & 90.67 & 88.95 \\
    \multicolumn{2}{c}{Fully supervision} & 88.53 & 88.53 & 88.53 & 88.53 & 94.87 & 94.87 & 94.87 & 94.87 & 96.15 & 96.15 & 96.15 & 96.15 \\ 
    \bottomrule

\end{tabular}
}
\label{tab:AL_combined}
\end{table*}

\subsection{Evaluation of FM-based CSAL for Classification}\label{subsec:cls_results}

\subsubsection{FM Comparison.}
We compare different FMs on the classification datasets in \textbf{Table~\ref{tab:foundation_models_combined}}, and for each FM and annotation ratio, we  present the average results obtained by the 7 sample selection methods. It can be observed that in average, the difference between these three types of feature extractors was very small, e.g., the accuracy ranged from 69.61\% to 71.56\% on the Derma dataset, and from 79.37\% to 81.44\% on the Breast dataset. In general, the families of DINO, CLIP and SAM obtained the best performance on the Derma, Pneumonia and Breast datasets, respectively, but the gap between the best and worst group is very small, i.e., $<0.5$\% in terms of accuracy. In addition, most FMs outperformed ResNet18 on the classification tasks, showing the advantage of FMs over  traditional simple pre-trained models for feature extraction in CSAL tasks. However, the average performance of these FMs was lower than that of ResNet18 on the Pneumonia dataset. 

\subsubsection{Comparison of Sample Selection Methods.}
\textbf{Table~\ref{tab:AL_combined}} shows a comparison of different sample selection methods on the classification datasets, where the average accuracy for each FM family is reported. When the SAM variants were used for feature extractor, only Coreset outperformed random selection on the Derma dataset, and FPS and Typiclust obtained the best performance on the Breast and Pneumonia datasets, respectively. When using DINO variants for feature extractor, Coreset and FPS could still maintain high performance, while RepDiv achieved the best performance on the Breast and Pneumonia datasets. In general, RepDiv obtained a leading performance in average on all the datasets with different feature extractors. Other methods such as FPS, Typiclust and ALPS also exhibit comparable performance.

\subsection{Discussion}

\noindent\textbf{Medical FM vs. General FMs.} In \textbf{Table~\ref{tab:foundation_models_heart_spleen}} and\textbf{~\ref{tab:foundation_models_kvasir_TN3K}}, comparison between MedCLIP and CLIP shows that despite the former is  better on Heart  dataset, it  did not consistently outperform general-purpose CLIP model in segmentation tasks and even obtained the worst performance on TN3K dataset. This is mainly because that the MedCLIP model was only pre-trained on X-ray images, which explains its good performance on X-ray image classification  (Pneumonia dataset), but poor results on the Derma and Breast datasets, as shown in \textbf{Table~\ref{tab:foundation_models_combined}}. This shows a higher generalizability of CLIP than MedCLIP to different medical image modalities. Results in \textbf{Table~\ref{tab:foundation_models_heart_spleen}}, \textbf{\ref{tab:foundation_models_kvasir_TN3K}} and \textbf{\ref{tab:foundation_models_combined}} also show that though MedSAM-ViT-B is better than SAM-VIT-B on the Heart dataset (averaged on 7 sample selection methods), it is inferior to the later on the Spleen, Kvasir, Breast and Pneumonia datasets. This demonstrates that current medical image FMs are not always better than general FMs in CSAL, possibly due to that the later has a more generalizable feature representation ability with a more diverse pre-training set.  Therefore, enhancing the diversity of the pre-training dataset is important for FMs when used in CSAL for medical image analysis. 

\noindent\textbf{Key Takeaway.} Based on the above analysis, we summarize our main findings below: 1) When FMs are used for feature extraction, most CSAL methods can obtain good performance in both segmentation and classification tasks, but different FMs show large variation of performance in CSAL, and random selection is still a strong competitor; 
2) For CSAL segmentation tasks, DINO series obtained the best performance in average, but there are minimal performance differences between SAM, CLIP and DINO families on CSAL classification tasks; 3) Medical FMs are not always superior to general FMs in CSAL for medical imaging; 4) Most FMs are better than simple ResNet18 pre-trained on ImageNet for feature extraction in CSAL, but the later may obtained the best performance on some segmentation datasets, necessitating more study on FM-based CSAL approaches. 5) Overall, ALPS, FPS, RepDiv, and Typiclust are robust for sample selection. ALPS yields the best average performance in segmentation tasks, while RepDiv excels in classification tasks, showing that different sample selection methods should be considered in different tasks. 

\noindent\textbf{Limitations and Future Work.}
Firstly, uncertainty-based sample selection methods are not considered in this work for comparison, which is mainly due to that using a pre-trained feature extractor is hard to obtain reliable task-relevant uncertainty estimation for downstream datasets in CSAL setting. Secondly, due to the lack of sufficient 3D foundation models, we only included 2D feature extractors for comparison, and it is  worthwhile to evaluate more FM-based feature extractors, especially those for 3D medical images.

\section{Conclusion}
We propose a novel FM-based CSAL paradigm for medical image analysis, and present a comprehensive  benchmark for evaluating the effectiveness of FMs for CSAL. We systematically compared 14 FMs on 7 medical image datasets with 6 different modalities for  classification and segmentation. The experimental results verified the general effectiveness of FMs in CSAL, where DINO family stands out as the most effective FM for CSAL in segmentation, and the difference between DINO, SAM and CLIP families are very small in classification tasks. In addition, we compared several sample selection methods, and find that ALPS takes the lead for segmentation tasks while RepDiv performs the best for classification. However, random selection remains a strong competitor. Our work sheds light on the potential, challenges and some limitations of FM-driven CSAL for medical imaging. 

\bibliography{aaai2026}

\begin{thebibliography}{45}
\providecommand{\natexlab}[1]{#1}

\bibitem[{Chen et~al.(2022)Chen, Bai, Huang, Lu, Wen, Yuille, and Zhou}]{chen2022making}
Chen, L.; Bai, Y.; Huang, S.; Lu, Y.; Wen, B.; Yuille, A.~L.; and Zhou, Z. 2022.
\newblock Making Your First Choice: To Address Cold Start Problem in Vision Active Learning.
\newblock arXiv:2210.02442.

\bibitem[{Deng et~al.(2009)Deng, Dong, Socher, Li, Li, and Fei-Fei}]{Deng2009ImageNet}
Deng, J.; Dong, W.; Socher, R.; Li, L.-J.; Li, K.; and Fei-Fei, L. 2009.
\newblock ImageNet: A large-scale hierarchical image database.
\newblock In \emph{2009 IEEE Conference on Computer Vision and Pattern Recognition}, 248--255.

\bibitem[{Doucet et~al.(2025)Doucet, Estermann, Aczel, and Wattenhofer}]{doucet2025bridging}
Doucet, P.; Estermann, B.; Aczel, T.; and Wattenhofer, R. 2025.
\newblock Bridging Diversity and Uncertainty in Active learning with Self-Supervised Pre-Training.
\newblock arXiv:2403.03728.

\bibitem[{Gong et~al.(2021)Gong, Chen, Wang, Xie, Mao, Yu, Chen, and Li}]{gong2021multi-task}
Gong, H.; Chen, G.; Wang, R.; Xie, X.; Mao, M.; Yu, Y.; Chen, F.; and Li, G. 2021.
\newblock Multi-Task Learning For Thyroid Nodule Segmentation With Thyroid Region Prior.
\newblock In \emph{2021 IEEE 18th International Symposium on Biomedical Imaging (ISBI)}, 257--261.

\bibitem[{Gong et~al.(2022)Gong, Chen, Chen, Li, Chen, and Li}]{gong2022thyroid}
Gong, H.; Chen, J.; Chen, G.; Li, H.; Chen, F.; and Li, G. 2022.
\newblock Thyroid Region Prior Guided Attention for Ultrasound Segmentation of Thyroid Nodules.
\newblock \emph{Computers in Biology and Medicine}, 106389: 1--12.

\bibitem[{Hacohen, Dekel, and Weinshall(2022)}]{hacohen2022active}
Hacohen, G.; Dekel, A.; and Weinshall, D. 2022.
\newblock Active Learning on a Budget: Opposite Strategies Suit High and Low Budgets.
\newblock In \emph{Proceedings of the 39th International Conference on Machine Learning}, volume 162 of \emph{Proceedings of Machine Learning Research}, 8175--8195.

\bibitem[{He et~al.(2016)He, Zhang, Ren, and Sun}]{He2016deep}
He, K.; Zhang, X.; Ren, S.; and Sun, J. 2016.
\newblock Deep Residual Learning for Image Recognition.
\newblock In \emph{2016 IEEE Conference on Computer Vision and Pattern Recognition (CVPR)}, 770--778.

\bibitem[{Holzmüller et~al.(2023)Holzmüller, Zaverkin, Kästner, and Steinwart}]{Holzmüller2023framework}
Holzmüller, D.; Zaverkin, V.; Kästner, J.; and Steinwart, I. 2023.
\newblock A Framework and Benchmark for Deep Batch Active Learning for Regression.
\newblock arXiv:2203.09410.

\bibitem[{Isensee et~al.(2021)Isensee, Jaeger, Kohl, Petersen, and {Maier-Hein}}]{isensee2021NnUNet}
Isensee, F.; Jaeger, P.~F.; Kohl, S. A.~A.; Petersen, J.; and {Maier-Hein}, K.~H. 2021.
\newblock {{nnU-Net}}: A Self-Configuring Method for Deep Learning-Based Biomedical Image Segmentation.
\newblock \emph{Nature Methods}, 18(2): 203--211.

\bibitem[{Jin et~al.(2025)Jin, Guo, Lin, Luo, and Chen}]{jin2025label}
Jin, C.; Guo, Z.; Lin, Y.; Luo, L.; and Chen, H. 2025.
\newblock Label-Efficient Deep Learning in Medical Image Analysis: Challenges and Future Directions.
\newblock arXiv:2303.12484.

\bibitem[{Jin et~al.(2022{\natexlab{a}})Jin, Yuan, Li, Wang, Wang, and Song}]{jin2022cold}
Jin, Q.; Yuan, M.; Li, S.; Wang, H.; Wang, M.; and Song, Z. 2022{\natexlab{a}}.
\newblock Cold-start active learning for image classification.
\newblock \emph{Information Sciences}, 616: 16--36.

\bibitem[{Jin et~al.(2022{\natexlab{b}})Jin, Yuan, Qiao, and Song}]{Jin2022Oneshot}
Jin, Q.; Yuan, M.; Qiao, Q.; and Song, Z. 2022{\natexlab{b}}.
\newblock One-Shot Active Learning for Image Segmentation via Contrastive Learning and Diversity-Based Sampling.
\newblock \emph{Knowledge-Based Systems}, 241: 108278.

\bibitem[{Jose et~al.(2024)Jose, Moutakanni, Kang, Baldassarre, Darcet, Xu, Li, Szafraniec, Ramamonjisoa, Oquab, Siméoni, Vo, Labatut, and Bojanowski}]{jose2024dinov2}
Jose, C.; Moutakanni, T.; Kang, D.; Baldassarre, F.; Darcet, T.; Xu, H.; Li, D.; Szafraniec, M.; Ramamonjisoa, M.; Oquab, M.; Siméoni, O.; Vo, H.~V.; Labatut, P.; and Bojanowski, P. 2024.
\newblock DINOv2 Meets Text: A Unified Framework for Image- and Pixel-Level Vision-Language Alignment.
\newblock arXiv:2412.16334.

\bibitem[{Kirillov et~al.(2023)Kirillov, Mintun, Ravi, Mao, Rolland, Gustafson, Xiao, Whitehead, Berg, Lo et~al.}]{kirillov2023segment}
Kirillov, A.; Mintun, E.; Ravi, N.; Mao, H.; Rolland, C.; Gustafson, L.; Xiao, T.; Whitehead, S.; Berg, A.~C.; Lo, W.-Y.; et~al. 2023.
\newblock Segment anything.
\newblock In \emph{Proceedings of the IEEE/CVF international conference on computer vision}, 4015--4026.

\bibitem[{Kumari and Singh(2023)}]{kumari2023data}
Kumari, S.; and Singh, P. 2023.
\newblock Data efficient deep learning for medical image analysis: A survey.
\newblock arXiv:2310.06557.

\bibitem[{Li et~al.(2024)Li, Chen, Yu, Liu, and Jia}]{Li2024BAL}
Li, J.; Chen, P.; Yu, S.; Liu, S.; and Jia, J. 2024.
\newblock BAL: Balancing Diversity and Novelty for Active Learning.
\newblock \emph{IEEE Transactions on Pattern Analysis and Machine Intelligence}, 46(5): 3653--3664.

\bibitem[{Liu et~al.(2023)Liu, Li, Yao, Fan, Hu, Dawant, Nath, Xu, and Oguz}]{Liu2023COLosSAL}
Liu, H.; Li, H.; Yao, X.; Fan, Y.; Hu, D.; Dawant, B.~M.; Nath, V.; Xu, Z.; and Oguz, I. 2023.
\newblock {{COLosSAL}}: A Benchmark for Cold-Start Active Learning for {{3D}} Medical Image Segmentation.
\newblock In \emph{Medical Image Computing and Computer Assisted Intervention -- {{MICCAI}} 2023}, 25--34.

\bibitem[{Loshchilov and Hutter(2019)}]{loshchilov2019decoupled}
Loshchilov, I.; and Hutter, F. 2019.
\newblock Decoupled Weight Decay Regularization.
\newblock arXiv:1711.05101.

\bibitem[{Luo et~al.(2024)Luo, Luo, Gao, and Wang}]{Luo2024Uncertainty}
Luo, Z.; Luo, X.; Gao, Z.; and Wang, G. 2024.
\newblock An Uncertainty-Guided Tiered Self-training Framework For Active Source-Free Domain Adaptation In Prostate Segmentation.
\newblock In \emph{Medical Image Computing and Computer Assisted Intervention -- MICCAI 2024}, 107--117. Cham.
\newblock ISBN 978-3-031-72114-4.

\bibitem[{Ma et~al.(2024)Ma, He, Li, Han, You, and Wang}]{ma2024segment}
Ma, J.; He, Y.; Li, F.; Han, L.; You, C.; and Wang, B. 2024.
\newblock Segment anything in medical images.
\newblock \emph{Nature Communications}, 15(1): 654.

\bibitem[{Ma et~al.(2025)Ma, Fu, Zhong, Zhu, and Wang}]{ma2025sugfw}
Ma, X.; Fu, J.; Zhong, L.; Zhu, N.; and Wang, G. 2025.
\newblock SUGFW: A SAM-Based Uncertainty-Guided Feature Weighting Framework for Cold Start Active Learning.
\newblock In \emph{International Conference on Medical Image Computing and Computer-Assisted Intervention}, 579--588. Springer.

\bibitem[{Moenning and Dodgson(2003)}]{moenning2003fast}
Moenning, C.; and Dodgson, N.~A. 2003.
\newblock Fast marching farthest point sampling.
\newblock Technical report, University of Cambridge, Computer Laboratory.

\bibitem[{Nath et~al.(2021)Nath, Yang, Landman, Xu, and Roth}]{Nath2021diminishing}
Nath, V.; Yang, D.; Landman, B.~A.; Xu, D.; and Roth, H.~R. 2021.
\newblock Diminishing Uncertainty Within the Training Pool: Active Learning for Medical Image Segmentation.
\newblock \emph{IEEE Transactions on Medical Imaging}, 40(10): 2534--2547.

\bibitem[{Oquab et~al.(2023)Oquab, Darcet, Moutakanni, Vo, Szafraniec, Khalidov, Fernandez, Haziza, Massa, El-Nouby et~al.}]{oquab2023dinov2}
Oquab, M.; Darcet, T.; Moutakanni, T.; Vo, H.; Szafraniec, M.; Khalidov, V.; Fernandez, P.; Haziza, D.; Massa, F.; El-Nouby, A.; et~al. 2023.
\newblock Dinov2: Learning robust visual features without supervision.
\newblock \emph{arXiv preprint arXiv:2304.07193}.

\bibitem[{Phan et~al.(2024)Phan, Xie, Qi, Liu, Liu, Zhang, Liao, Wu, To, and Verjans}]{Phan2024decomposing}
Phan, V. M.~H.; Xie, Y.; Qi, Y.; Liu, L.; Liu, L.; Zhang, B.; Liao, Z.; Wu, Q.; To, M.-S.; and Verjans, J.~W. 2024.
\newblock Decomposing Disease Descriptions for Enhanced Pathology Detection: A Multi-Aspect Vision-Language Pre-Training Framework.
\newblock In \emph{2024 IEEE/CVF Conference on Computer Vision and Pattern Recognition (CVPR)}, 11492--11501.

\bibitem[{Pogorelov et~al.(2017)Pogorelov, Randel, Griwodz, Eskeland, de~Lange, Johansen, Spampinato, Dang-Nguyen, Lux, Schmidt, Riegler, and Halvorsen}]{Pogorelov2017KVASIR}
Pogorelov, K.; Randel, K.~R.; Griwodz, C.; Eskeland, S.~L.; de~Lange, T.; Johansen, D.; Spampinato, C.; Dang-Nguyen, D.-T.; Lux, M.; Schmidt, P.~T.; Riegler, M.; and Halvorsen, P. 2017.
\newblock KVASIR: A Multi-Class Image Dataset for Computer Aided Gastrointestinal Disease Detection.
\newblock In \emph{Proceedings of the 8th ACM on Multimedia Systems Conference}, MMSys'17, 164--169.
\newblock ISBN 978-1-4503-5002-0.

\bibitem[{Radford et~al.(2021)Radford, Kim, Hallacy, Ramesh, Goh, Agarwal, Sastry, Askell, Mishkin, Clark et~al.}]{radford2021learning}
Radford, A.; Kim, J.~W.; Hallacy, C.; Ramesh, A.; Goh, G.; Agarwal, S.; Sastry, G.; Askell, A.; Mishkin, P.; Clark, J.; et~al. 2021.
\newblock Learning transferable visual models from natural language supervision.
\newblock In \emph{International conference on machine learning}, 8748--8763.

\bibitem[{Ravi et~al.(2024)Ravi, Gabeur, Hu, Hu, Ryali, Ma, Khedr, R{\"a}dle, Rolland, Gustafson et~al.}]{ravi2024sam}
Ravi, N.; Gabeur, V.; Hu, Y.-T.; Hu, R.; Ryali, C.; Ma, T.; Khedr, H.; R{\"a}dle, R.; Rolland, C.; Gustafson, L.; et~al. 2024.
\newblock Sam 2: Segment anything in images and videos.
\newblock \emph{arXiv preprint arXiv:2408.00714}.

\bibitem[{Rokuss et~al.(2025)Rokuss, Kirchhoff, Akbal, Kovacs, Roy, Ulrich, Wald, Rotkopf, Schlemmer, and Maier-Hein}]{rokuss2025lesionlocator}
Rokuss, M.; Kirchhoff, Y.; Akbal, S.; Kovacs, B.; Roy, S.; Ulrich, C.; Wald, T.; Rotkopf, L.~T.; Schlemmer, H.-P.; and Maier-Hein, K. 2025.
\newblock LesionLocator: Zero-Shot Universal Tumor Segmentation and Tracking in 3D Whole-Body Imaging.
\newblock arXiv:2502.20985.

\bibitem[{Sener and Savarese(2017)}]{sener2017active}
Sener, O.; and Savarese, S. 2017.
\newblock Active learning for convolutional neural networks: A core-set approach.
\newblock \emph{arXiv preprint arXiv:1708.00489}.

\bibitem[{Shen et~al.(2020)Shen, Tian, Dong, Zhang, Yan, Che, Yao, Luo, and Han}]{Shen2020deep}
Shen, H.; Tian, K.; Dong, P.; Zhang, J.; Yan, K.; Che, S.; Yao, J.; Luo, P.; and Han, X. 2020.
\newblock Deep Active Learning for Breast Cancer Segmentation on Immunohistochemistry Images.
\newblock In \emph{Medical Image Computing and Computer Assisted Intervention -- MICCAI 2020}, 509--518.

\bibitem[{Simpson et~al.(2019)Simpson, Antonelli, Bakas, Bilello, Farahani, van Ginneken, Kopp-Schneider, Landman, Litjens, Menze, Ronneberger, Summers, Bilic, Christ, Do, Gollub, Golia-Pernicka, Heckers, Jarnagin, McHugo, Napel, Vorontsov, Maier-Hein, and Cardoso}]{simpson2019large}
Simpson, A.~L.; Antonelli, M.; Bakas, S.; Bilello, M.; Farahani, K.; van Ginneken, B.; Kopp-Schneider, A.; Landman, B.~A.; Litjens, G.; Menze, B.; Ronneberger, O.; Summers, R.~M.; Bilic, P.; Christ, P.~F.; Do, R. K.~G.; Gollub, M.; Golia-Pernicka, J.; Heckers, S.~H.; Jarnagin, W.~R.; McHugo, M.~K.; Napel, S.; Vorontsov, E.; Maier-Hein, L.; and Cardoso, M.~J. 2019.
\newblock A large annotated medical image dataset for the development and evaluation of segmentation algorithms.
\newblock arXiv:1902.09063.

\bibitem[{Wang et~al.(2024)Wang, Jin, Li, Liu, Wang, and Song}]{wang2024comprehensive}
Wang, H.; Jin, Q.; Li, S.; Liu, S.; Wang, M.; and Song, Z. 2024.
\newblock A comprehensive survey on deep active learning in medical image analysis.
\newblock arXiv:2310.14230.

\bibitem[{Wang, Lian, and Yu(2022)}]{Wang2022Unsupervised}
Wang, X.; Lian, L.; and Yu, S.~X. 2022.
\newblock Unsupervised Selective Labeling for More Effective Semi-Supervised Learning.
\newblock In \emph{Computer Vision -- {{ECCV}} 2022}, 427--445.

\bibitem[{Wang et~al.(2022)Wang, Wu, Agarwal, and Sun}]{wang2022medclip}
Wang, Z.; Wu, Z.; Agarwal, D.; and Sun, J. 2022.
\newblock Medclip: Contrastive learning from unpaired medical images and text.
\newblock In \emph{Proceedings of the Conference on Empirical Methods in Natural Language Processing. Conference on Empirical Methods in Natural Language Processing}, volume 2022, 3876.

\bibitem[{Wei, Lin, and Caesar(2024)}]{Wei2024BaSAL}
Wei, J.; Lin, Y.; and Caesar, H. 2024.
\newblock BaSAL: Size-Balanced Warm Start Active Learning for LiDAR Semantic Segmentation.
\newblock In \emph{2024 IEEE International Conference on Robotics and Automation (ICRA)}, 18258--18264.

\bibitem[{Werner et~al.(2024)Werner, Burchert, Stubbemann, and Schmidt-Thieme}]{Werner2024cross-domain}
Werner, T.; Burchert, J.; Stubbemann, M.; and Schmidt-Thieme, L. 2024.
\newblock A Cross-Domain Benchmark for Active Learning.
\newblock arXiv:2408.00426.

\bibitem[{Yang et~al.(2023)Yang, Shi, Wei, Liu, Zhao, Ke, Pfister, and Ni}]{Yang2023MedMNIST}
Yang, J.; Shi, R.; Wei, D.; Liu, Z.; Zhao, L.; Ke, B.; Pfister, H.; and Ni, B. 2023.
\newblock MedMNIST v2 - A large-scale lightweight benchmark for 2D and 3D biomedical image classification.
\newblock \emph{Scientific Data}, 10(1).

\bibitem[{Yang et~al.(2017)Yang, Zhang, Chen, Zhang, and Chen}]{Yang2017Suggestive}
Yang, L.; Zhang, Y.; Chen, J.; Zhang, S.; and Chen, D.~Z. 2017.
\newblock Suggestive Annotation: A Deep Active Learning Framework for Biomedical Image Segmentation.
\newblock In \emph{Medical Image Computing and Computer Assisted Intervention - MICCAI 2017}, 399--407.

\bibitem[{Yehuda et~al.(2022)Yehuda, Dekel, Hacohen, and Weinshall}]{Yehuda2022Probcover}
Yehuda, O.; Dekel, A.; Hacohen, G.; and Weinshall, D. 2022.
\newblock Active Learning Through a Covering Lens.
\newblock In Koyejo, S.; Mohamed, S.; Agarwal, A.; Belgrave, D.; Cho, K.; and Oh, A., eds., \emph{Advances in Neural Information Processing Systems}, volume~35, 22354--22367.

\bibitem[{Yuan, Lin, and Boyd-Graber(2020)}]{yuan2020cold}
Yuan, M.; Lin, H.-T.; and Boyd-Graber, J. 2020.
\newblock Cold-start Active Learning through Self-supervised Language Modeling.
\newblock In \emph{Proceedings of the 2020 Conference on Empirical Methods in Natural Language Processing (EMNLP)}, 7935--7948.

\bibitem[{Zhan et~al.(2021)Zhan, Liu, Li, and Chan}]{Zhan2021comparative}
Zhan, X.; Liu, H.; Li, Q.; and Chan, A. 2021.
\newblock A Comparative Survey: Benchmarking for Pool-based Active Learning.
\newblock In \emph{Proceedings of the 30th International Joint Conference on Artificial Intelligence, IJCAI 2021}, IJCAI International Joint Conference on Artificial Intelligence, 4679--4686.

\bibitem[{Zhang et~al.(2022)Zhang, Li, Liu, Zhang, Su, Zhu, Ni, and Shum}]{zhang2022dino}
Zhang, H.; Li, F.; Liu, S.; Zhang, L.; Su, H.; Zhu, J.; Ni, L.~M.; and Shum, H.-Y. 2022.
\newblock Dino: Detr with improved denoising anchor boxes for end-to-end object detection.
\newblock \emph{arXiv preprint arXiv:2203.03605}.

\bibitem[{Zhong et~al.(2025)Zhong, Qian, Liao, Huang, Liu, Zhang, and Wang}]{Zhong2025UniSAL}
Zhong, L.; Qian, K.; Liao, X.; Huang, Z.; Liu, Y.; Zhang, S.; and Wang, G. 2025.
\newblock UniSAL: Unified Semi-supervised Active Learning for histopathological image classification.
\newblock \emph{Medical Image Analysis}, 102: 103542.

\bibitem[{Zhu et~al.(2025)Zhu, Ye, Zhong, Yue, Zhang, and Wang}]{zhu2025csal}
Zhu, N.; Ye, P.; Zhong, L.; Yue, Q.; Zhang, S.; and Wang, G. 2025.
\newblock CSAL-3D: Cold-Start Active Learning for 3D Medical Image Segmentation via SSL-Driven Uncertainty-Reinforced Diversity Sampling.
\newblock In \emph{International Conference on Medical Image Computing and Computer-Assisted Intervention}, 120--130. Springer.

\end{thebibliography}

\newpage
\onecolumn
\section{Appendix}
In this technical appendix, we provide additional details for experimental details. First, we provide a visualization of selected samples of different CSAL methods to better illustrate their performance difference. Secondly, we present the raw data for all experiments to ensure transparency.

\subsection{Visualization of Selected Samples of Different CSAL Methods}\label{sec:vis_samples}
\begin{figure}[h]  
  \centering
  \includegraphics[width=0.85\linewidth]{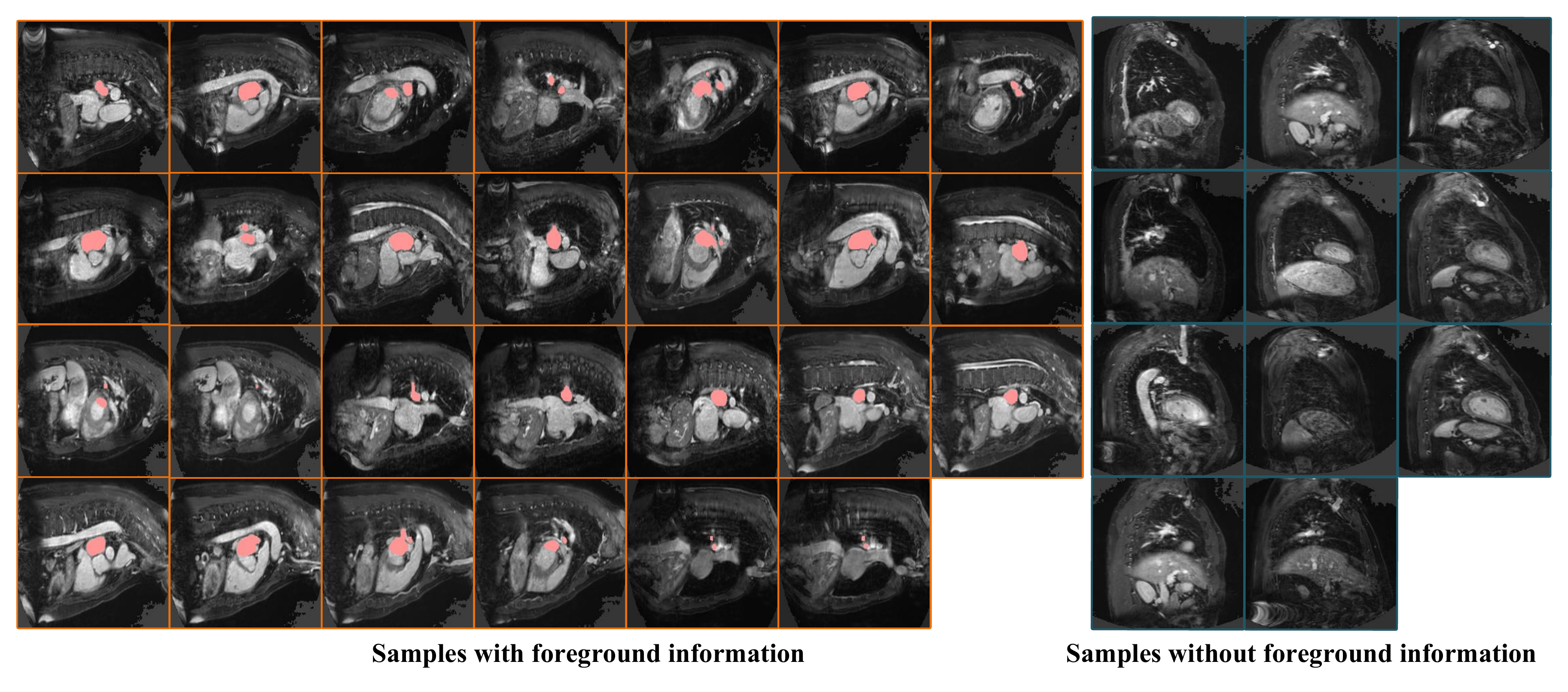}    
  \caption{The selected samples using MedCLIP-ViT as FM feature extractor and ALPS as the sample selection method for Heart dataset (left:samples with foreground information, right: samples without foreground information).}
  \label{fig:heart_alps}
\end{figure}

\begin{figure}[h]  
  \centering
  \includegraphics[width=0.85\linewidth]{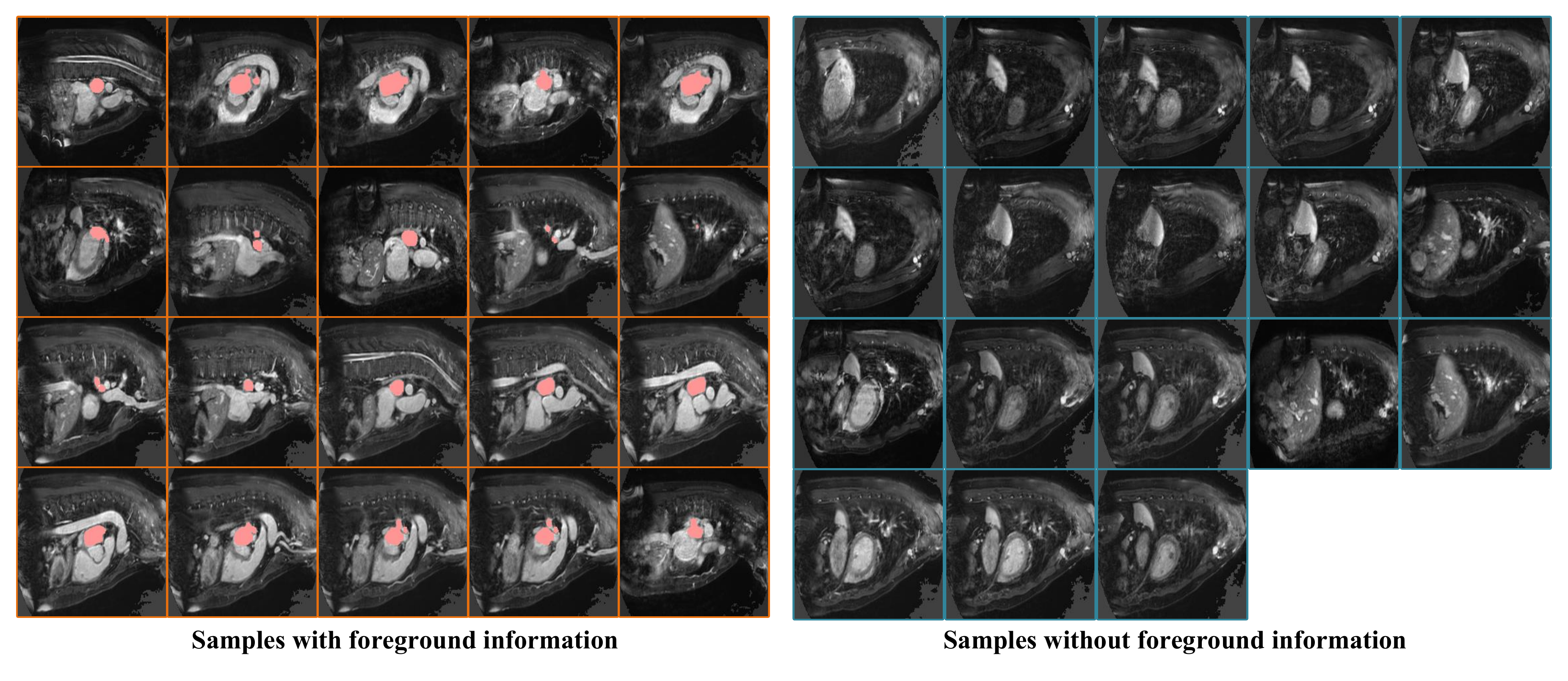}    
  \caption{The selected samples using MedCLIP-ViT as FM feature extractor and BAL as the sample selection method for Heart dataset (left:samples with foreground information, right: samples without foreground information).}
  \label{fig:heart_bal}
\end{figure}

\begin{figure}[h]  
  \centering
  \includegraphics[width=0.82\linewidth]{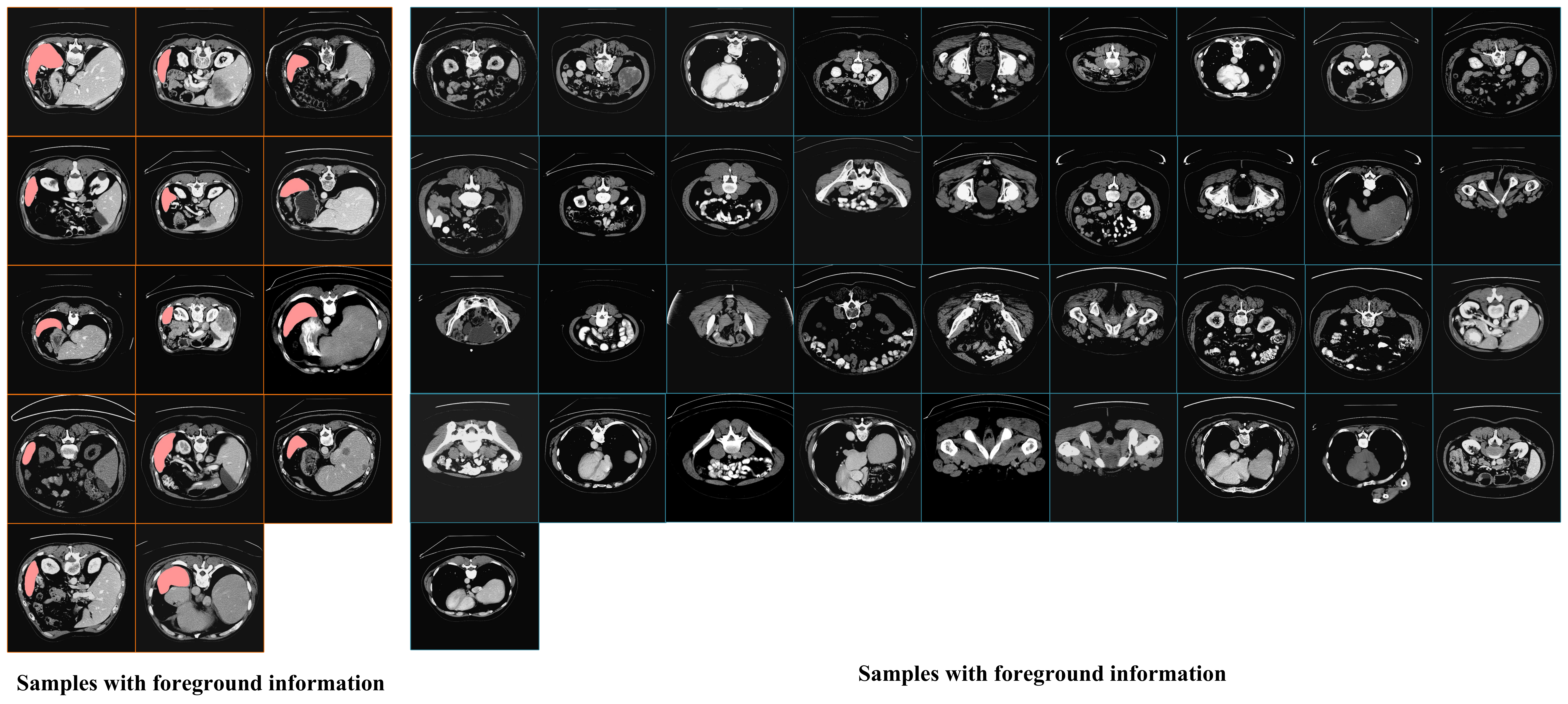}    
  \caption{The selected samples using MedCLIP-ViT as FM feature extractor and ALPS as the sample selection method for Spleen dataset (left:samples with foreground information, right: samples without foreground information).}
  \label{fig:spleen_alps}
\end{figure}

\begin{figure}[h]  
  \centering
  \includegraphics[width=0.82\linewidth]{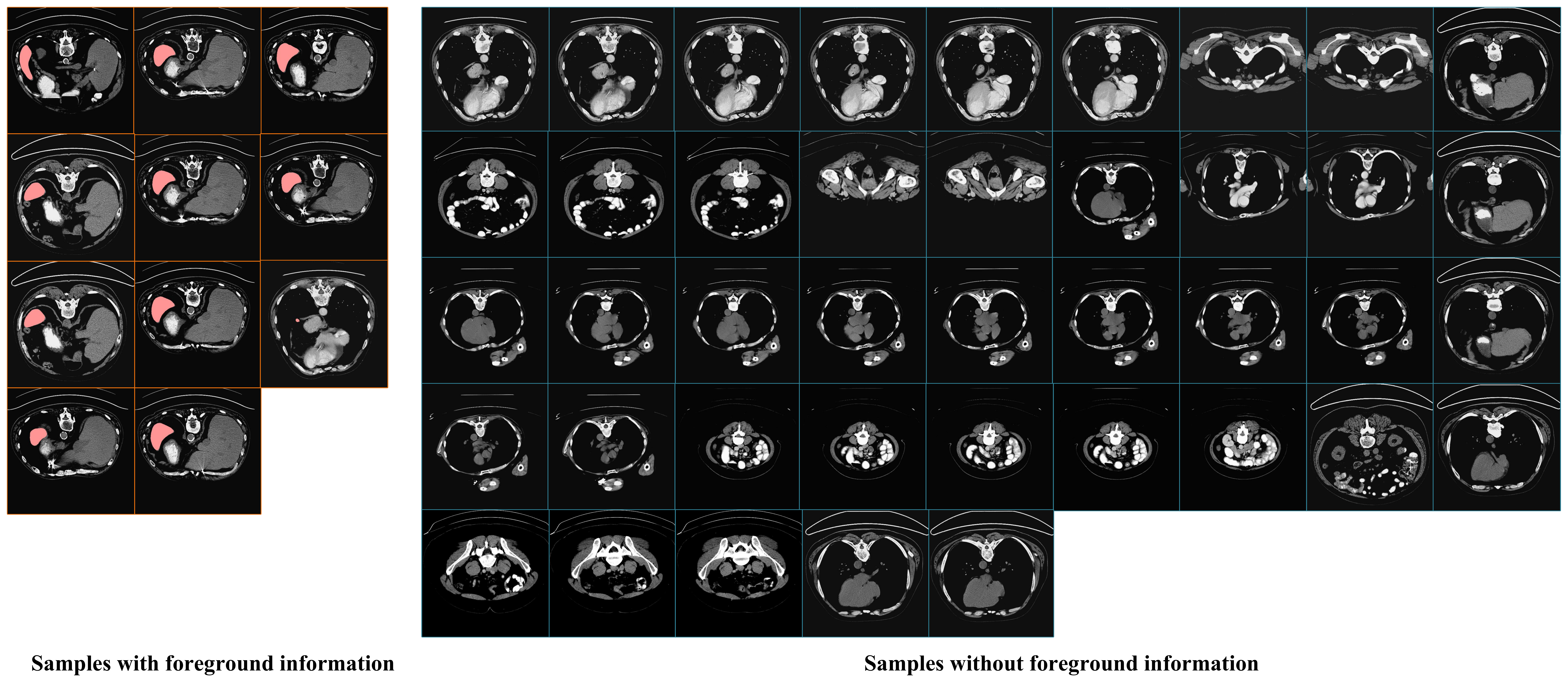}    
  \caption{The selected samples using MedCLIP-ViT as FM feature extractor and BAL as the sample selection method for Spleen dataset (left:samples with foreground information, right: samples without foreground information). }
  \label{fig:spleen_bal}
\end{figure}

We visualize the selected data samples using MedCLIP-ViT from ALPS and BAL with the annotation budget of 38 slices (5\%) for Heart dataset in \textbf{Fig.~\ref{fig:heart_alps}} and in \textbf{Fig.~\ref{fig:heart_bal}}, and annotation budget of 51 slices (2\%) for Spleen dataset in \textbf{Fig.~\ref{fig:spleen_alps}} and in \textbf{Fig.~\ref{fig:spleen_bal}}. It can be seen that BAL prefers to choose samples near the cluster boundaries, and 2D slices with a small foreground region can easily be present in such regions in the feature space.  If certain CSAL method selects too few samples containing foreground information, it can significantly hinder the training of the segmentation model. In extreme cases, this may even result in the inability to obtain meaningful segmentation outcomes (e.g., Dice = 0). 

\subsection{Raw Data}\label{sec:raw_data}
In this section, we present the raw experimental data for all FM-CSAL paired evaluation, also with the ResNet18 competitor. We first provide raw segmentation results on \textbf{Table~\ref{tab:raw_heart}} for Heart dataset, \textbf{Table~\ref{tab:raw_spleen}} for Spleen dataset, and \textbf{Table~\ref{tab:raw_TN3K_Kvasir}} for Kvasir and TN3K dataset. For 3D segmentation task, we report the Dice score and HD95. For 2D segmentation task, we report Dice score. Following segmentation results, we then provide raw classification results in accuracy (\%) on  \textbf{Table~\ref{tab:raw_cls}}. 

\vspace{5pt}

\noindent\textbf{Remarks.} It can be noticed that for 3D segmentation task (Heart in \textbf{Table~\ref{tab:raw_heart}} and Spleen in \textbf{Table~\ref{tab:raw_spleen}}), there exist some experimental results which do not yield any meaningful result (with 0\% testing Dice score and infinite HD95) since certain CSAL methods select too many 2D slices with no or few foreground information.

\onecolumn
\subsubsection{Segmentation Results on Heart Dataset}
\begin{longtable}{cccccccc}
\caption{Dice (\%) and HD95 (voxel) results on Heart} \label{tab:addlabel} \\
\toprule
\multicolumn{1}{c}{\multirow{2}{*}{Foundation Model}} & \multicolumn{1}{c}{\multirow{2}{*}{AL Method}} & \multicolumn{2}{c}{5\%} & \multicolumn{2}{c}{10\%} & \multicolumn{2}{c}{15\%} \\
\cmidrule(lr){3-4} \cmidrule(lr){5-6} \cmidrule(lr){7-8}
 & & Dice & HD95 & Dice & HD95 & Dice & HD95 \\
\midrule
\endfirsthead

\multicolumn{8}{c}%
{{\bfseries \tablename\ \thetable{} -- continued from previous page}} \\
\toprule
\multicolumn{1}{c}{\multirow{2}{*}{Foundation Model}} & \multicolumn{1}{c}{\multirow{2}{*}{AL Methods}} & \multicolumn{2}{c}{5\%} & \multicolumn{2}{c}{10\%} & \multicolumn{2}{c}{15\%} \\
\cmidrule(lr){3-4} \cmidrule(lr){5-6} \cmidrule(lr){7-8}
 & & Dice & HD95 & Dice & HD95 & Dice & HD95 \\
\midrule
\endhead

\bottomrule
\multicolumn{8}{r}{{Continued on next page}} \\
\endfoot

\bottomrule
\endlastfoot

    —— & Random1 & 65.44 \textsuperscript{4.68} & 19.17 \textsuperscript{9.22} & 78.52 \textsuperscript{2.75} & 11.42 \textsuperscript{6.81} & 79.50 \textsuperscript{2.47} & 12.27 \textsuperscript{7.43} \\
    —— & Random2 & 69.42 \textsuperscript{4.26} & 10.46 \textsuperscript{2.31} & 75.87 \textsuperscript{3.86} & 8.52 \textsuperscript{1.53} & 81.40 \textsuperscript{2.71} & 6.53 \textsuperscript{1.13} \\
    —— & Random3 & 64.70 \textsuperscript{5.58} & 19.46 \textsuperscript{9.39} & 77.64 \textsuperscript{2.80} & 7.39 \textsuperscript{1.15} & 79.07 \textsuperscript{7.01} & 7.74 \textsuperscript{3.88} \\
    —— & Random4 & 69.77 \textsuperscript{3.28} & 13.16 \textsuperscript{0.85} & 77.89 \textsuperscript{2.99} & 9.86 \textsuperscript{2.46} & 80.63 \textsuperscript{2.87} & 7.02 \textsuperscript{1.49} \\
    —— & Random5 & 66.74 \textsuperscript{11.14} & 14.41 \textsuperscript{6.65} & 77.79 \textsuperscript{4.27} & 8.82 \textsuperscript{3.53} & 79.39 \textsuperscript{4.94} & 8.08 \textsuperscript{4.69} \\
    ResNet18 & ALPS  & 74.04 \textsuperscript{6.62} & 9.70 \textsuperscript{3.67} & 81.10 \textsuperscript{3.83} & 7.06 \textsuperscript{2.80} & 80.61 \textsuperscript{2.49} & 6.47 \textsuperscript{1.20} \\
    ResNet18 & Coreset & 68.18 \textsuperscript{4.37} & 9.83 \textsuperscript{2.08} & 79.99 \textsuperscript{4.66} & 6.69 \textsuperscript{2.51} & 81.62 \textsuperscript{3.70} & 6.25 \textsuperscript{1.78} \\
    ResNet18 & FPS   & 74.49 \textsuperscript{5.91} & 10.60 \textsuperscript{5.22} & 80.09 \textsuperscript{2.52} & 6.51 \textsuperscript{1.30} & 80.63 \textsuperscript{3.68} & 6.36 \textsuperscript{2.07} \\
\end{longtable}

\onecolumn
\subsubsection{Segmentation Results on Spleen Dataset}
\begin{longtable}{cccccccc}
\caption{Dice (\%) and HD95 (voxel) results on Spleen} \label{tab:raw_spleen} \\
\toprule
\multicolumn{1}{c}{\multirow{2}{*}{Foundation Model}} & \multicolumn{1}{c}{\multirow{2}{*}{AL Method}} & \multicolumn{2}{c}{2\%} & \multicolumn{2}{c}{4\%} & \multicolumn{2}{c}{10\%} \\
\cmidrule(lr){3-4} \cmidrule(lr){5-6} \cmidrule(lr){7-8}
 & & Dice & HD95 & Dice & HD95 & Dice & HD95 \\
\midrule
\endfirsthead

\multicolumn{8}{c}%
{{\bfseries \tablename\ \thetable{} -- continued from previous page}} \\
\toprule
\multicolumn{1}{c}{\multirow{2}{*}{Foundation Model}} & \multicolumn{1}{c}{\multirow{2}{*}{AL Methods}} & \multicolumn{2}{c}{2\%} & \multicolumn{2}{c}{4\%} & \multicolumn{2}{c}{10\%} \\
\cmidrule(lr){3-4} \cmidrule(lr){5-6} \cmidrule(lr){7-8}
 & & Dice & HD95 & Dice & HD95 & Dice & HD95 \\
\midrule
\endhead

\bottomrule
\multicolumn{8}{r}{{Continued on next page}} \\
\endfoot

\bottomrule
\endlastfoot

    ——    & Random1 & 52.28 \textsuperscript{30.76} & 101.54 \textsuperscript{109.50} & 63.96 \textsuperscript{23.54} & 95.46 \textsuperscript{97.95} & 70.21 \textsuperscript{17.66} & 62.11 \textsuperscript{78.53} \\
    ——    & Random2 & 53.95 \textsuperscript{27.03} & 96.99 \textsuperscript{102.62} & 62.70 \textsuperscript{14.81} & 99.55 \textsuperscript{103.39} & 67.32 \textsuperscript{22.95} & 53.50 \textsuperscript{80.13} \\
    ——    & Random3 & 43.32 \textsuperscript{15.62} & 217.12 \textsuperscript{80.11} & 58.43 \textsuperscript{16.45} & 144.84 \textsuperscript{115.71} & 62.51 \textsuperscript{23.72} & 19.09 \textsuperscript{17.34} \\
    ——    & Random4 & 66.39 \textsuperscript{19.08} & 100.11 \textsuperscript{108.07} & 76.14 \textsuperscript{10.46} & 91.52 \textsuperscript{98.31} & 67.96 \textsuperscript{23.04} & 21.94 \textsuperscript{24.52} \\
    ——    & Random5 & 48.49 \textsuperscript{7.82} & 211.33 \textsuperscript{42.52} & 62.64 \textsuperscript{17.92} & 134.54 \textsuperscript{109.51} & 67.42 \textsuperscript{23.04} & 14.60 \textsuperscript{13.57} \\

    ResNet18 & ALPS  & 49.41 \textsuperscript{27.21} & 32.09 \textsuperscript{28.88} & 54.67 \textsuperscript{16.95} & 54.51 \textsuperscript{74.34} & 70.67 \textsuperscript{20.94} & 48.77 \textsuperscript{76.20} \\
    ResNet18 & Coreset & 61.73 \textsuperscript{23.94} & 112.10 \textsuperscript{107.61} & 65.26 \textsuperscript{24.74} & 16.05 \textsuperscript{20.51} & 71.39 \textsuperscript{21.73} & 43.95 \textsuperscript{64.58} \\
    ResNet18 & FPS   & 59.26 \textsuperscript{25.21} & 93.22 \textsuperscript{99.50} & 54.50 \textsuperscript{27.41} & 72.32 \textsuperscript{93.06} & 75.02 \textsuperscript{20.38} & 30.86 \textsuperscript{45.68} \\
\end{longtable}

\onecolumn
\subsubsection{Segmentation Results on Kvasir and TN3K Datasets}
\begin{longtable}{ccccc|ccc}
\caption{Dice (\%) results on Kvasir and TN3K} \label{tab:raw_TN3K_Kvasir} \\
\toprule
% \centering
\multicolumn{1}{c}{\multirow{2}{*}{Foundation Model}} & \multicolumn{1}{c}{\multirow{2}{*}{AL Method}} & \multicolumn{3}{c}{Kvasir} & \multicolumn{3}{c}{TN3K} \\
\cmidrule(lr){3-5} \cmidrule(lr){6-8}
 & & 10\% & 15\% & 20\% & 5\% & 7.5\% & 10\% \\
\midrule
\endfirsthead

\multicolumn{8}{c}%
{{\bfseries \tablename\ \thetable{} -- continued from previous page}} \\
\toprule
\multicolumn{1}{c}{\multirow{2}{*}{Foundation Model}} & \multicolumn{1}{c}{\multirow{2}{*}{AL Methods}} & \multicolumn{3}{c}{Kvasir} & \multicolumn{3}{c}{TN3K} \\
\cmidrule(lr){3-5} \cmidrule(lr){6-8}
  & & 10\% & 15\% & 20\% & 5\% & 7.5\% & 10\% \\
\midrule
\endhead

\bottomrule
\multicolumn{8}{r}{{Continued on next page}} \\
\endfoot

\bottomrule
\endlastfoot

    ——    & Random1 & 65.82 \textsuperscript{29.90} & 71.90 \textsuperscript{28.79} & 73.15 \textsuperscript{27.63} & 67.98 \textsuperscript{26.46} & 70.44 \textsuperscript{26.31} & 72.18 \textsuperscript{25.72} \\
    ——    & Random2 & 66.73 \textsuperscript{29.56} & 71.96 \textsuperscript{26.78} & 76.47 \textsuperscript{25.54} & 63.50 \textsuperscript{29.37} & 67.05 \textsuperscript{25.62} & 68.62 \textsuperscript{27.96} \\
    ——    & Random3 & 66.88 \textsuperscript{30.55} & 72.77 \textsuperscript{28.41} & 75.59 \textsuperscript{26.06} & 64.31 \textsuperscript{29.45} & 66.40 \textsuperscript{28.31} & 71.06 \textsuperscript{25.53} \\
    ——    & Random4 & 70.06 \textsuperscript{28.10} & 73.89 \textsuperscript{28.61} & 74.62 \textsuperscript{26.91} & 63.03 \textsuperscript{27.83} & 69.39 \textsuperscript{27.03} & 71.69 \textsuperscript{25.97} \\
    ——    & Random5 & 67.23 \textsuperscript{29.97} & 70.90 \textsuperscript{29.34} & 74.73 \textsuperscript{27.34} & 65.62 \textsuperscript{28.27} & 70.25 \textsuperscript{25.64} & 70.87 \textsuperscript{25.39} \\

    ResNet18 & ALPS  & 69.69 \textsuperscript{26.65} & 72.93 \textsuperscript{25.40} & 74.86 \textsuperscript{26.20} & 65.68 \textsuperscript{26.87} & 68.11 \textsuperscript{27.70} & 69.81 \textsuperscript{27.63} \\
    ResNet18 & Coreset & 66.92 \textsuperscript{28.56} & 75.27 \textsuperscript{23.30} & 77.64 \textsuperscript{22.57} & 59.51 \textsuperscript{32.38} & 63.56 \textsuperscript{32.15} & 66.84 \textsuperscript{28.39} \\
    ResNet18 & FPS   & 66.56 \textsuperscript{28.46} & 71.31 \textsuperscript{27.44} & 78.89 \textsuperscript{22.84} & 59.27 \textsuperscript{30.72} & 68.23 \textsuperscript{27.45} & 70.64 \textsuperscript{27.14} \\
\end{longtable}

\onecolumn
\subsubsection{Classification Results on Derma, Breast and Pneumonia Datasets}
\begin{longtable}{ccccc|ccc|ccc}
\caption{Accuracy (\%) results on classification datasets} \label{tab:raw_cls} \\
\toprule
\multicolumn{1}{c}{\multirow{2}{*}{Foundation Model}} & \multicolumn{1}{c}{\multirow{2}{*}{AL Methods}} & \multicolumn{3}{c}{Derma} & \multicolumn{3}{c}{Breast} & \multicolumn{3}{c}{Pneumonia} \\
\cmidrule(lr){3-5} \cmidrule(lr){6-8} \cmidrule(lr){9-11}
& & 1\% & 2\% & 3\% & 5\% & 10\% & 15\% & 1\% & 2\% & 3\% \\
\midrule
\endfirsthead

\multicolumn{11}{c}{{\bfseries \tablename\ \thetable{} -- continued from previous page}} \\
\toprule
\multicolumn{1}{c}{\multirow{2}{*}{Foundation Model}} & \multicolumn{1}{c}{\multirow{2}{*}{AL Methods}} & \multicolumn{3}{c}{Derma} & \multicolumn{3}{c}{Breast} & \multicolumn{3}{c}{Pneumonia} \\
\cmidrule(lr){3-5} \cmidrule(lr){6-8} \cmidrule(lr){9-11}
& & 1\% & 2\% & 3\% & 5\% & 10\% & 15\% & 1\% & 2\% & 3\% \\
\midrule
\endhead

\midrule
\multicolumn{11}{r}{{Continued on next page}} \\
\endfoot

\bottomrule
\endlastfoot

    ——    & Random1 & 69.53 & 71.92 & 72.82 & 76.28 & 82.69 & 81.41 & 81.25 & 84.13 & 87.66 \\
    ——    & Random2 & 68.48 & 72.12 & 74.96 & 72.44 & 81.41 & 82.69 & 93.43 & 92.95 & 94.07 \\
    ——    & Random3 & 69.78 & 69.53 & 71.92 & 77.56 & 77.56 & 85.90 & 84.78 & 90.71 & 90.06 \\
    ——    & Random4 & 70.67 & 71.22 & 73.22 & 67.95 & 79.49 & 85.90 & 90.38 & 92.47 & 93.27 \\
    ——    & Random5 & 69.43 & 69.03 & 72.57 & 74.36 & 79.49 & 86.54 & 78.21 & 92.63 & 88.30 \\

    ResNet18 & ALPS  & 67.03 & 68.73 & 74.02 & 80.77 & 83.97 & 79.49 & 88.30 & 93.43 & 92.63 \\
    ResNet18 & Coreset & 68.03 & 69.23 & 71.07 & 73.08 & 77.56 & 80.13 & 89.90 & 90.71 & 91.03 \\
    ResNet18 & FPS   & 69.08 & 70.97 & 73.82 & 76.92 & 77.56 & 85.26 & 85.90 & 91.19 & 87.82 \\
\end{longtable} 

% \bibliographystyle{unsrtnat}
% \bibliography{references_new}  %%% Uncomment this line and comment out the ``thebibliography'' section below to use the external .bib file (using bibtex) .

\end{document}